\documentclass[format=acmsmall, review=false, screen=true]{acmart}

\usepackage{threeparttable}

\usepackage{booktabs} 

\usepackage[ruled]{algorithm2e} 

\SetAlFnt{\small}
\SetAlCapFnt{\small}
\SetAlCapNameFnt{\small}
\SetAlCapHSkip{0pt}
\IncMargin{-\parindent}

\acmJournal{TIST}
\acmVolume{0}
\acmNumber{0}
\acmArticle{0}
\acmYear{0}
\acmMonth{0}
\copyrightyear{0}
\acmArticleSeq{0}

\setcopyright{acmlicensed}

\begin{document}
\title[A Review of Co-saliency Detection] {A Review of Co-saliency Detection Algorithms: Fundamentals, Applications, and Challenges}

\author{Dingwen Zhang}
\orcid{1234-5678-9012-3456}
\affiliation{%
  \institution{Northwestern Polytechnical University}
  \city{Xi'an}
  \state{Shaanxi}
  \postcode{710129}
  \country{China}}
\author{Huazhu Fu}
\affiliation{%
  \institution{Institute for Infocomm Research, Agency for Science, Technology and Research}
  \city{Singapore}
}
\author{Junwei Han}
\affiliation{%
 \institution{Northwestern Polytechnical University}
  \city{Xi'an}
  \state{Shaanxi}
  \postcode{710129}
  \country{China}}
\author{Ali Borji}
\affiliation{%
  \institution{University of Central Florida}
  \city{Orlando}
  \state{Florida}
  \country{USA}
}
\author{Xuelong Li}
\affiliation{%
  \institution{Xi'an Institute of Optics and Precision Mechanics, Chinese Academy of Sciences}
  \city{Xi'an}
  \state{Shaanxi}
  \postcode{710129}
  \country{China}}

\begin{abstract}
Co-saliency detection is a newly emerging and rapidly growing research area in computer vision community. As a novel branch of visual saliency, co-saliency detection refers to the discovery of common and salient foregrounds from two or more relevant images, and can be widely used in many computer vision tasks. The existing co-saliency detection algorithms mainly consist of three components: extracting effective features to represent the image regions, exploring the informative cues or factors to characterize co-saliency, and designing effective computational frameworks to formulate co-saliency. Although numerous methods have been developed, the literature is still lacking a deep review and evaluation of co-saliency detection techniques. In this paper, we aim at providing a comprehensive review of the fundamentals, challenges, and applications of co-saliency detection. Specifically, we provide an overview of some related computer vision works, review the history of co-saliency detection, summarize and categorize the major algorithms in this research area, discuss some open issues in this area, present the potential applications of co-saliency detection, and finally point out some unsolved challenges and promising future works. We expect this review to be beneficial to both fresh and senior researchers in this field, and give insights to researchers in other related areas regarding the utility of co-saliency detection algorithms.
\end{abstract}

%
%


%
%


\thanks{This work was supported in part by the National Science Foundation of China under Grants 61522207 and 61473231, in part by the Doctorate Foundation through Northwestern Polytechnical University, and in part by the Excellent Doctorate Foundation through Northwestern Polytechnical University.

 Authors' addresses: D.~Zhang and J.~Han (corresponding author), School of Automation, Northwestern Polytechnical University, Xi'an, 710072, China; H.~Fu the Ocular Imaging Department, Institute for Infocomm Research, Agency for Science, Technology and Research, Singapore; A. Borji Center for Research in Computer Vision, University of Central Florida , Orlando, USA; X.~Li, Center for OPTical IMagery Analysis and Learning, State Key Laboratory of Transient Optics and Photonics, Xi'an Institute of Optics and Precision Mechanics, Chinese Academy of Sciences, Xi'an, China.}

\maketitle

\renewcommand{\shortauthors}{D. Zhang et al.}

\section{Introduction}

\begin{figure}[h]
\centerline{\includegraphics[width=8cm]{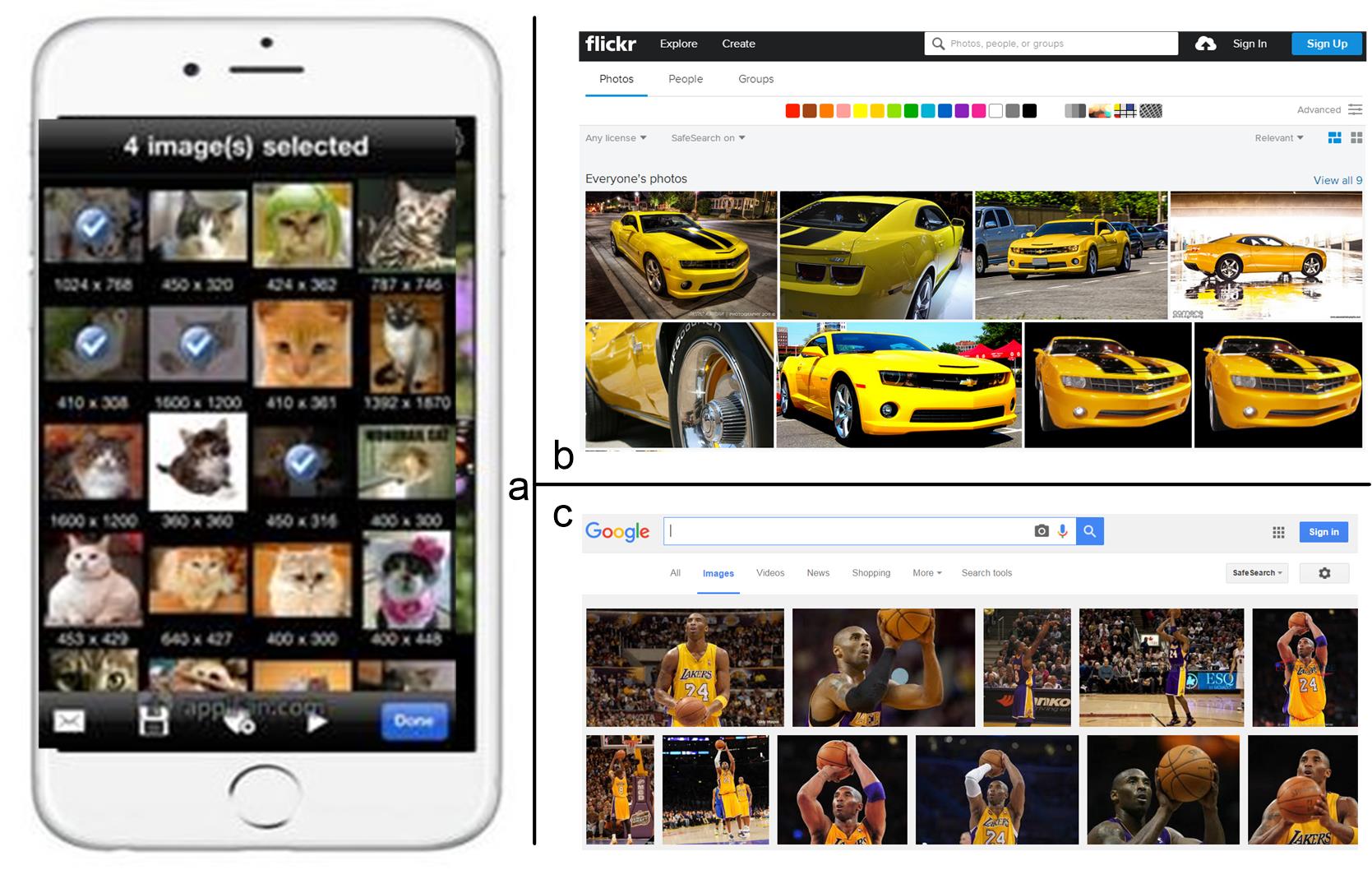}}
\caption{Massive related image data can be collected from (a) smart phones, (b) the image sharing Web, and (c) image search engines. The co-occurred patterns can be utilized to represent the main content of a group of images.}
\label{img_cover}
\end{figure}

Many imaging equipment, such as digital cameras and smart phones, are able to acquire large collections of image or video data. Photo-sharing websites, such as Flickr and Facebook, have also made such data accessible. These are illustrated in Fig.~\ref{img_cover}. Consequently, nowadays people are more likely to face a large number of images, that are typically huge in size and share common objects or events. Compared to individual images, a group of images contains richer and more useful information. Within the image group, the frequently occurred patterns or the prime foregrounds can be utilized to represent the main content of the image group. The co-saliency task is thus derived with the goal of establishing effective computational systems to endow such capabilities to machines.

\begin{figure}[t]
\centerline{\includegraphics[width=8cm]{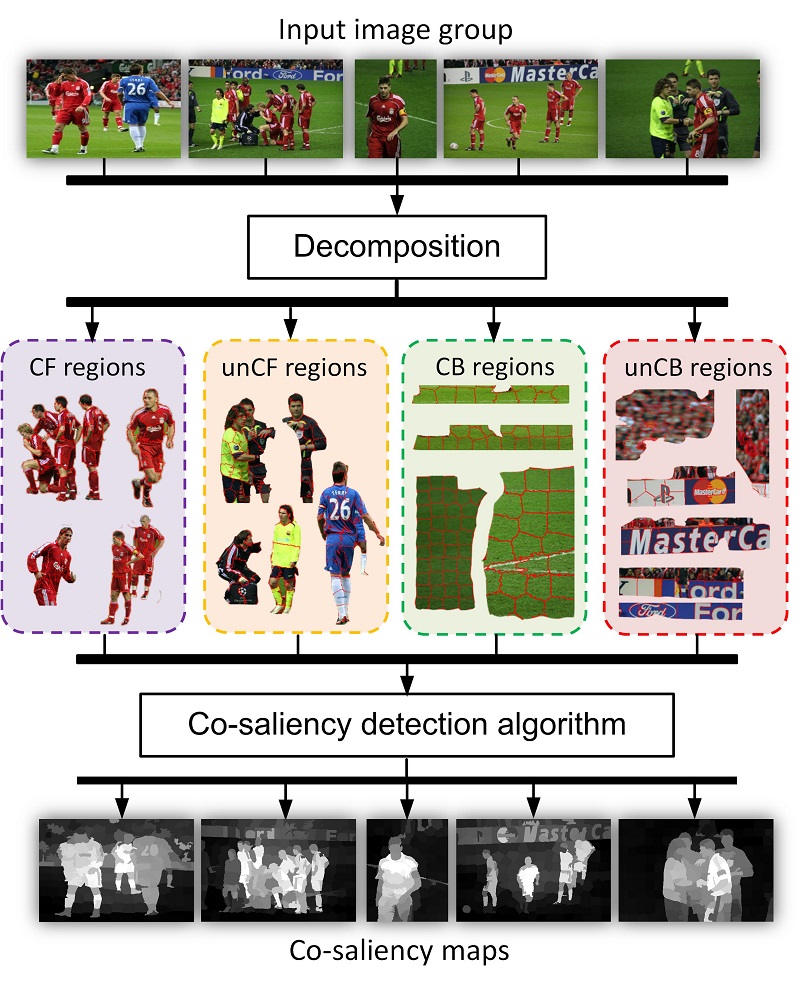}}
\caption{An image set can be separated into the common foreground (CF), common background (CB), uncommon foreground (unCF), and uncommon background (unCB) regions. The task of co-saliency detection is to generate a co-saliency map highlighting the common foreground (e.g., the players in red cloth).}
\label{img_comp}
\end{figure}

 {Co-saliency indicates the common and salient visual stimulus residing in a given image group. Co-saliency detection is a computational problem which aims at highlighting the common and salient foreground regions in an image group. Notice that as the desired semantic categories of co-salient objects are unknown, such information is dependent on the specific content of the given image group and needs to be inferred by the designed algorithm.} Specifically, given an image set, an image group can be divided into four components: the common foreground (CF), common background (CB), uncommon foreground (unCF), and uncommon background (unCB) regions, as shown in Fig.~\ref{img_comp}. The task of co-saliency detection is to discover the common and salient regions (i.e., common foregrounds) from the multiple images. In order to discover the real CF regions from the noisy CB, unCF, and unCB regions, the existing co-saliency detection approaches mainly focus on solving three key problems: 1) extracting representative features to describe the image foregrounds, 2) exploring the informative cues or factors to characterize co-saliency, and 3) designing effective computational frameworks to formulate co-saliency.

Regarding the feature representation, a large number of co-saliency detection methods mainly use low-level features~\cite{Chang2011,Fu2013,Tan2013}, such as color histograms, Gabor filters, or SIFT descriptors. These methods assume that the co-salient objects in multiple related images should share certain low-level consistency. Another group of methods, e.g.,~\cite{Chen2010,Li2013,Li2011,Cao_TIP_2014}, additionally use mid-level features to detect co-saliency. These methods usually consider the prediction results of the previous (co-)saliency detection algorithms as the mid-level features and seek to explore the consistency of the co-salient objects based on them. More recently, some high-level semantic features\footnote{ {Generally, low-level feature denote the pixel-level representation. The middle-level feature means the saliency submaps generated by the existing (co-)saliency methods. The high-level features, such as~\cite{han2015object,zeiler2011adaptive,li2014object}, refer to the features that can provide semantic information of the data. In this paper, high-level features mainly (but not only) indicate deep layers of a CNN.}} were proposed to be used in some co-saliency detection approaches~\cite{Zhang_CVPR_2015,Zhang_ICCV_2015,Zhang_TNNLS_2015}. These approaches assume that the co-salient objects appearing in multiple images should share stronger consistency in high-level concepts.

The most frequently used informative cues to characterize co-saliency include intra- and inter- image similarities. Specifically, the former is used to ensure that the detected object regions are salient in each individual image, while the latter is used to ensure that the detected object regions are commonly appearing in a given image group. Apparently,  these two factors are designed to reflect the basic definition of co-saliency (i.e. `salient' and `common') as mentioned above. In addition to these two factors, some other cues, such as the objectness~\cite{Li2011,li_ICME_2014,Liu_SPL_2014,Zhang_TNNLS_2015}, the center prior~\cite{Fu2013,Chen_ICPR_2014}, and the border connectivity \cite{Ye_SPL_2015}, have also been utilized in co-saliency detection. Moreover, some works have also introduced the inter-group separability to better discriminate the common foregrounds and background~\cite{Zhang_CVPR_2015,Zhang_ICCV_2015,Ge201669}.

From the existing literature, we observe that the trend is changing from the bottom-up frameworks to the top-down ones. Most of the early co-saliency detection methods were bottom-up (e.g.,~\cite{Fu2013,Li2013,Li_SPL_2015,li_ICME_2014}). They exploited features or computational metrics that were relied heavily on the human knowledge for extracting salient objects and interpretation of the results. With the rapid development of new machine learning algorithms, some co-saliency detection frameworks have been proposed that are  designed in a top-down manner. For example, Cao et al.~\cite{Cao_TIP_2014} and Zhang et al.~\cite{Zhang_ICCV_2015} adopted the fusion-based and learning-based approaches, respectively, to detect co-saliency based on the top-down priors inferred in a given image group.

In the past several years, co-saliency detection field has gained extensive interest. Different from the traditional saliency detection on a single image~\cite{borji2013state,Borji2015_TIP}, co-saliency detection aims at discovering the common and salient foregrounds from an image group containing two or more relevant images. Notice that the categories, intrinsic characteristics, and locations of these objects are often unknown. Due to its superior scalability, co-saliency detection has been widely adopted in several computer vision tasks, such as image/video co-segmentation~\cite{fu2015TIP,Fu2015_CVPR,Wang2015TIP,Tao2017AAAI}, object co-localization~\cite{tang2014coLO,Joulin2014,Cho_2015_CVPR}, and weakly supervised learning~\cite{Siva2011,Siva2013CVPR}. Thus, it is of great interest to review the newly developed co-saliency detection techniques and provide a comprehensive discussion of the fundamentals that exist in the major co-saliency detection methods. It is also important to enumerate the existing and potential applications that can benefit from co-saliency detection techniques, and discuss the unsolved challenges in this research area that need to be addressed in the future.

In this paper, we provide a comprehensive review of the recent progress in this area, which could help: 1) the fresh researchers in learning the key techniques and concepts in this area and incorporating them in their research, 2) the senior researchers in identifying and addressing the unsolved problems to build more effective and efficient systems, and 3) the researchers working on other relevant areas in exploring the possibilities of benefiting from co-saliency detection techniques.

The rest of this paper is organized as follows. Section~\ref{sec_related} provides an overview of some other computer vision tasks related to co-saliency detection. Section~\ref{sec_chronology} provides a brief history of co-saliency detection. The major co-saliency detection algorithms are summarized in Section~\ref{sec_algorithm}. We mention the evaluation methodologies in Section~\ref{sec-evaluation}
followed by applications of co-saliency techniques in Section~\ref{sec_application}. Section
~\ref{sec_discussion} discusses some important matters and challenges regarding saliency detection. Finally, conclusions and future works are proposed in Section 9 .~\ref{sec_conclusion}.

\section{Related Areas}
\label{sec_related}

\begin{table*} [!t]
	\centering
    \tiny
	\caption{Comparison of co-saliency detection and its related areas.}
	\begin{tabular}{p{1.8cm}p{1.6cm}p{1.5cm}p{2.8cm}p{1.5cm}p{2.5cm}}
		\hline
		{}           & Saliency detection &  Co-segmentation  & Weakly supervised localization &  Video saliency  &  Co-saliency  \\ \hline \hline
		Basic model  &  Visual attention  &     Common pattern      &            Category            & Visual attention &    Computational co-attention    \\
		Computation scale &       Single       &        Multiple         &            Multiple            &      Video       &         Multiple          \\
		Target       &   Salient region   & Common objects &   Category-dependence object   &  Motion object   & Salient and common object \\
		Output       &  Probability map   &       Binary mask       &          Bounding box          & Probability map  &      Probability map      \\
\hline
	\end{tabular}%
	\label{Tab_Related2}%
\end{table*}%

In this section, we discuss some related areas to co-saliency detection including saliency object detection, object co-segmentation, weakly supervised localization, and video saliency. A brief summary is shown in Table~\ref{Tab_Related2}.

\subsection{Saliency detection}

Saliency detection aims at highlighting regions that attract the human visual attention in a single image. Generally,  saliency detection models are inspired by the human visual attention mechanisms, and can be divided into two categories: eye fixation prediction models~\cite{han2015two,liu2015predicting} and salient object detection models~\cite{borji2016reconciling,Borji2015_TIP,lu2013real}. The former attempts to predict fixation locations when people are freely viewing natural scenes, while the latter attempts to detect and segment the entire extent of salient objects that pop out from their contexts and attract people's attention. To predict salient locations, some early methods were mainly based on local contrast~\cite{itti1998model,Liu2011b,han2015two} and global contrast~\cite{cheng2015global} with the assumption that salient regions in each image should be distinctive from their surroundings or the entire image context. Later on, some methods started to introduce other factors such as background prior~\cite{wei2012geodesic,Zhu_2014_CVPR}, high-level priors~\cite{borji2012boosting,Goferman2012}, and depth information~\cite{Lang2012,Cheng2014,Peng2014}. More recently, deep neural networks have been adopted for saliency detection (e.g.,~\cite{han2014background,liu2015predicting,Huang_2015_ICCV,chen2016disc,xia2016bottom}) and have achieved remarkable performance.

\begin{figure}[h]
\centerline{\includegraphics[width=8cm]{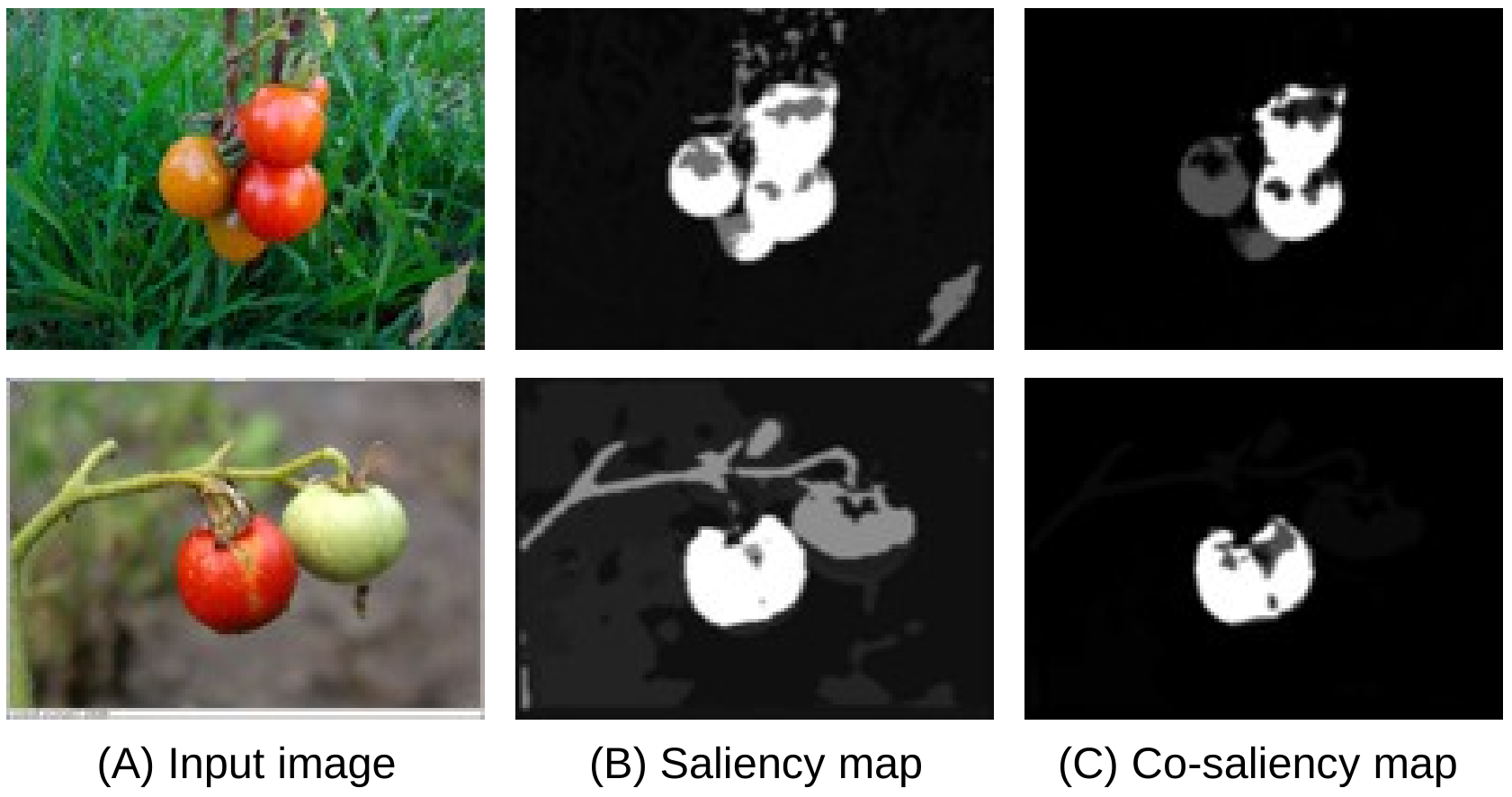}}
\caption{ {An example of the difference between saliency and co-saliency. In contrast to saliency detection (B) for each single image, the co-saliency detection (C) needs to highlight the common salient foregrounds (e.g., red fruit) from multiple related images, i.e., the image pair in this example.}}
\label{img_saliency}
\end{figure}

 {Rather than modeling the human visual attention mechanisms during free-viewing of single images, co-saliency makes effort in identifying informative and salient regions in a group of related images (see Table~\ref{Tab_Related2}).}  Specifically, different from saliency detection, co-saliency is not only influenced by the contrast factor within each individual image, but also determined by the consistent co-occurring patterns among the multiple related images, as shown in Fig.~\ref{img_saliency}. Thus, co-saliency detection is a relatively newer trend than the saliency detection. It is worthwhile to mention that most of the co-saliency detection methods can be employed for detecting saliency in a single image simply by setting the number of images to one.
 
\subsection{ {Video Saliency}}
\label{sec_spatialtem_saliency}

 {Given a video sequence, video saliency~\cite{mahadevan2010spatiotemporal,fang2014video,li2013temporally}, a.k.a  spatiotemporal saliency, aims at combining both spatial and temporal information cues to infer salient object regions in video frames or shorts~\cite{nguyen2013static}. The key issue here is how to incorporate informative motion cues that are crucial in video processing. Specifically, inspired by biological mechanisms of motion-based perceptual grouping, Mahadevan et al.~\cite{mahadevan2010spatiotemporal} proposed a discriminant formulation of center-surround saliency by modeling the spatiotemporal video patches as dynamic textures. Their method offers a principled joint characterization of the spatial and temporal components of saliency. Fang et al.~\cite{fang2014video} first generated the spatial and temporal saliency maps separately and then merged them into one by a spatiotemporally adaptive entropy-based uncertainty weighting approach. To explore the short-term continuity, Li et al.~\cite{li2013temporally} proposed a unified regional framework which considered both intra-frame saliency and inter-frame saliency based on the multiple features including color and motion.}

 {As can be seen, both video saliency and co-saliency need to explore the spatial saliency to identify the high contrast image regions within each single image/frame. However, in contrast to the spatial saliency, video saliency needs to additionally explore the temporal motion cues residing in adjacent video frames, whereas co-saliency detection needs to further explore the correspondence or consistency among the images from the image group.}

\subsection{Co-segmentation}
\label{sec_related_coseg}

Co-segmentation techniques aim to assign multiple labels to segment both the common \textit{things} and \textit{stuffs} rather than just separating the common foreground objects from the background regions~\cite{Rother2006,Hochbaum2009,45joulin2010discriminative,47Kim2011distributed,Fu2014CVPR}. More recently, by observing that in most applications of co-segmentation, the regions of interest are objects (\textit{things}) rather than \textit{stuffs} \footnote{In this context, \textit{things} usually refers to the objects like cars and cows while \textit{stuffs} usually refers to the texture material like grass and rocks~\cite{48joulin2012multi,forsyth1996finding}. }, Vicente et al.~defined the object co-segmentation task more specifically and applied the objectness measure in their formulation explicitly~\cite{51vicente2011object}. Wang et al.~proposed a semi-supervised learning algorithm to exploit the similarity from both the limited training image foregrounds and the common objects shared among the unlabeled images~\cite{52wang2013semi}.

According to our observation, the difference of such two research topics mainly lies in three-fold aspects: 1) Co-saliency detection focuses on discovering the common and salient objects from the given image groups while co-segmentation method additionally tends to  segment out all the similar regions including both foreground and background~\cite{48joulin2012multi,47Kim2011distributed}, 2) Co-segmentation usually needs semi- or interactive- supervision~\cite{52wang2013semi,batra2010icoseg} (where some object regions need to be labeled in advance) while co-saliency detection is implemented in an unsupervised or weakly supervised manner, and 3) Compared with co-segmentation, co-saliency usually needs to introduce common pattern analysis into the contrast-based visual attention mechanisms. Although different, these two research areas are strong related as mentioned in~\cite{Chang2011,Fu2015_CVPR,fu2015TIP,Tao2017AAAI}. Co-saliency detection can be applied to provide the useful prior of the common foregrounds for the co-segmentation task.

\subsection{Weakly supervised localization}
\label{sec_related_wsl}

Weakly supervised localization (WSL) is another research topic closely related to co-saliency detection. The basic goal of WSL is to jointly localize common objects from the given weakly labeled images and learn their appearance models~\cite{Galleguillos2008,Nguyen2009,Deselaers2012}. However, this actually leads to a chicken-and-egg scenario because in order to localize the common objects, one needs to enable models to capture their appearance. To learn such appearance models, one needs to localize the object instances in advance. In practice, most WSL methods~\cite{Pandey2011,Cinbis2014,Wang2014} choose to find ways to initialize the object instances at first and then jointly refine the appearance model of the common objects and the localization annotations. For example, Siva et al.~\cite{Siva2011} proposed to make use of the objectness, intra-image similarity, and inter-image variance to initialize the common object instances. Shi et al.~\cite{shi2012transfer} proposed to transfer the mapping relationship between the selection accuracy and the appearance similarity from an auxiliary fully annotated dataset to make an initialization. Zhang et al.~\cite{zhang2015weakly} and Han et al.~\cite{han2015object} also adopted the saliency cue in their initialization stages. Song et al.~\cite{song2014weakly} proposed to discover the frequent configurations of discriminative visual patterns for the initialization.

As can be seen, although no effort has been made to adopt co-saliency detection in WSL, the co-saliency detection problem still appears to have a close relationship with the problems in WSL, especially for localizing the common objects from the given weakly labeled images. Eventually, co-saliency detection can be conveniently applied to improve the initialization performance of the common object instances in WSL. This because without additional prior knowledge or supervision, ``salient'' and ``common'' are the two useful pieces of information for initializing the WSL frameworks and they are what co-saliency detection models seek to figure out. The difference between co-saliency detection and WSL is also obvious: WSL needs to additionally learn the appearance models of the target category via incorporating more top-down factors and its final output is the bounding boxes with the corresponding confidence scores rather than the probability map output generated by the co-saliency detection algorithms.

\begin{table*}[t]
	\centering
    \tiny
	\caption{ {Some co-saliency detection algorithms. (L), (M), and (H) indicate the low-, mid-, and high-level feature, respectively.}}
	\begin{tabular}{p{1cm}p{1cm}p{1cm}p{0.2\columnwidth}p{0.05\columnwidth}p{0.3\columnwidth}p{0.07\columnwidth}}
		\hline
		Name                                 & Year & Pub.  & Feature                                                           & Img.\# & Method                                                                                                                                                                            & Dataset                                            \\ \hline\hline
		Jacob~\cite{Jacobs2010}              & 2010 & UIST  & Single saliency map (M), Nearest Neighbor error / incoherence (L) & 2      & Fusion-based method: fusing multi-features via SVM                                                                                                                                      & None                                               \\
		Chen~\cite{Chen2010}                 & 2010 & ICIP  & Sparse feature (M)                                                & 2      & Bottom-up method: minimizing the distribution divergence on co-salient objects via KL-divergence                                                                                  & None                                               \\
		Li~\cite{Li2011}                     & 2011 & TIP   & Single image saliency maps (M), color and texture (L)             & 2      & Fusion-based method: weighted combination of single-image saliency and inter-image similarity                                                                                     & Image pair                                         \\
		Chang~\cite{Chang2011}               & 2011 & CVPR  & SIFT (L)                                                          & many   & Bottom-up method: generating the co-saliency based on the saliency and repeatedness                                                                                               & MSRC                                               \\
		Fu~\cite{Fu2013}                     & 2013 & TIP   & Pixel location (L), color (L),  Gabor filter (L)                  & many   & Bottom-up method: employing contrast, spatial, corresponding cues in cluster-level                                                                                                & Image pair,  iCoseg                                 \\
		Li~\cite{Li2013}                     & 2013 & TMM   & Pyramid representation (M), and co-occurrence descriptor (M)      & many   & Fusion-based method: weighted combination of intra-image and inter-image saliency                                                                                                 & Image pair, iCoseg, MSRC                           \\
		Cao~\cite{Cao_TIP_2014}              & 2014 & TIP   & Saliency submap (M)                                                         & many   & Fusion-based method: combining multiple saliency maps via low rank matrix recovery                                                                                                  & Image pair, iCoseg                                 \\
		Liu~\cite{Liu_SPL_2014}              & 2014 & SPL   & Color (L)                                                         & many   & Fusion-based method: fusing intra-saliency, object prior*, global similarity via multi-level segmentation**                                                                          & Image pair, iCoseg                                 \\
		Ye~\cite{Ye_SPL_2015}                & 2015 & SPL   & Color (L), SIFT (L)                                               & many   & Bottom-up method: recovering co-salient object from pre-defined exemplars                                                                                                         & iCoseg, MSRC                                       \\
		Zhang~\cite{Zhang_TNNLS_2015}        & 2015 & TNNLS & Color (L), steerable pyramid filter (L), Gabor filter (L)         & many   & Bottom-up method: exploring intra-saliency prior via cross-domain transfer and mining deep inter-saliency via stacked denoising autoencoder                                       & Image pair, iCoseg                                 \\
		Du~\cite{Du_SPL_2015}                & 2015 & SPL   & Color (L)                                                         & many   & Learning-based method: determining inter-saliency from all the generated groups via nearest neighbor searching                                                                    & Internet dataset~\cite{rubinstein2013unsupervised} \\
		Zhang~\cite{Zhang_2016_IJCV}         & 2016 & IJCV  & Semantic feature (H)                                              & many   & Bottom-up method: integrating deep semantic feature and wide inter-group heterogeneity to explore the intra?image contrast, intra?group consistency, and inter?group seperability & iCoseg, MSRC, Cosal2015                            \\
		Zhang~\cite{zhang2016PAMI}           & 2016 & TPAMI & Deep feature (H)                                                  & many   & Learning-based method: self-learning without hand-crafted metrics based on SP-MIL                                                                                                 & iCoseg, MSRC                                       \\
		Jerri~\cite{Jerripothula2016} & 2016 & ECCV  & GIST (L), SIFT (L), Color (L)                                     & many   & Fusion-based method: video co-saliency based on inter-video, intra-video, and motion co-saliency maps                                                                             & YouTube  Dataset~\cite{prest2012learning}          \\
		Song~\cite{Song2016}                 & 2016 & SPL   & Multiple Feature Bagging (M)                                      & many   & Bottom-up method: utilizing feature bagging to randomly generate multiple clustering results and the cluster-level co-saliency maps                                               & RGBD coseg dataset~\cite{Fu2015_CVPR}              \\ \hline
	\end{tabular}%
	\label{Tab_list}%

\begin{tablenotes}
\item[*] *Object prior is the prior knowledge used to reflect which image regions are more likely to be objects rather than backgrounds.

\item[**] **Global similarity refers to the similarity of a certain image region to image regions in other images from the given image group.

\end{tablenotes}

\end{table*}%

\section{A Brief Chronology} 
\label{sec_chronology}

The history of co-saliency detection can be traced back to 2010, when Jacobs et al. first defined the concept of co-saliency as calculating the importance, or saliency, of image pixels in the context of other related images~\cite{Jacobs2010}. After that, the interest started with the goal of discovering co-saliency from image pairs~\cite{Chen2010,Li2011,Tan2013}.  To extend co-saliency to image groups with more than two related images, the work in~\cite{Fu2013} made use of multiple co-saliency cues (i.e., the contrast cue, spatial cue, and corresponding cue) to detect the co-salient regions. More recently, Zhang et al.~\cite{Zhang_CVPR_2015,Zhang_ICCV_2015,Zhang_TNNLS_2015} introduced the widely used deep learning techniques~\cite{brahma2015deep} for co-saliency detection.

Some major co-saliency detection algorithms are listed in Table~\ref{Tab_list}\footnote{ {A more complete list of co-saliency papers can be found in our project page: \url{http://hzfu.github.io/proj_cosal_review.html}}}. Even though the history of co-saliency detection techniques only spans several years, it has already gone through encouraging development and has shown promising prospects for the future. Firstly, the accuracy of the co-saliency detection techniques has improved dramatically. Secondly, the development in terms of the quality and quantity of publications, the ideas and frameworks proposed have also increased during the past few years.

\section{Major Algorithms}
\label{sec_algorithm}

The strategies of co-saliency detection can be grouped into three main categories: bottom-up methods, fusion-based methods, and learning-based methods.

\begin{figure*}[t]
\centerline{\includegraphics[width=15 cm]{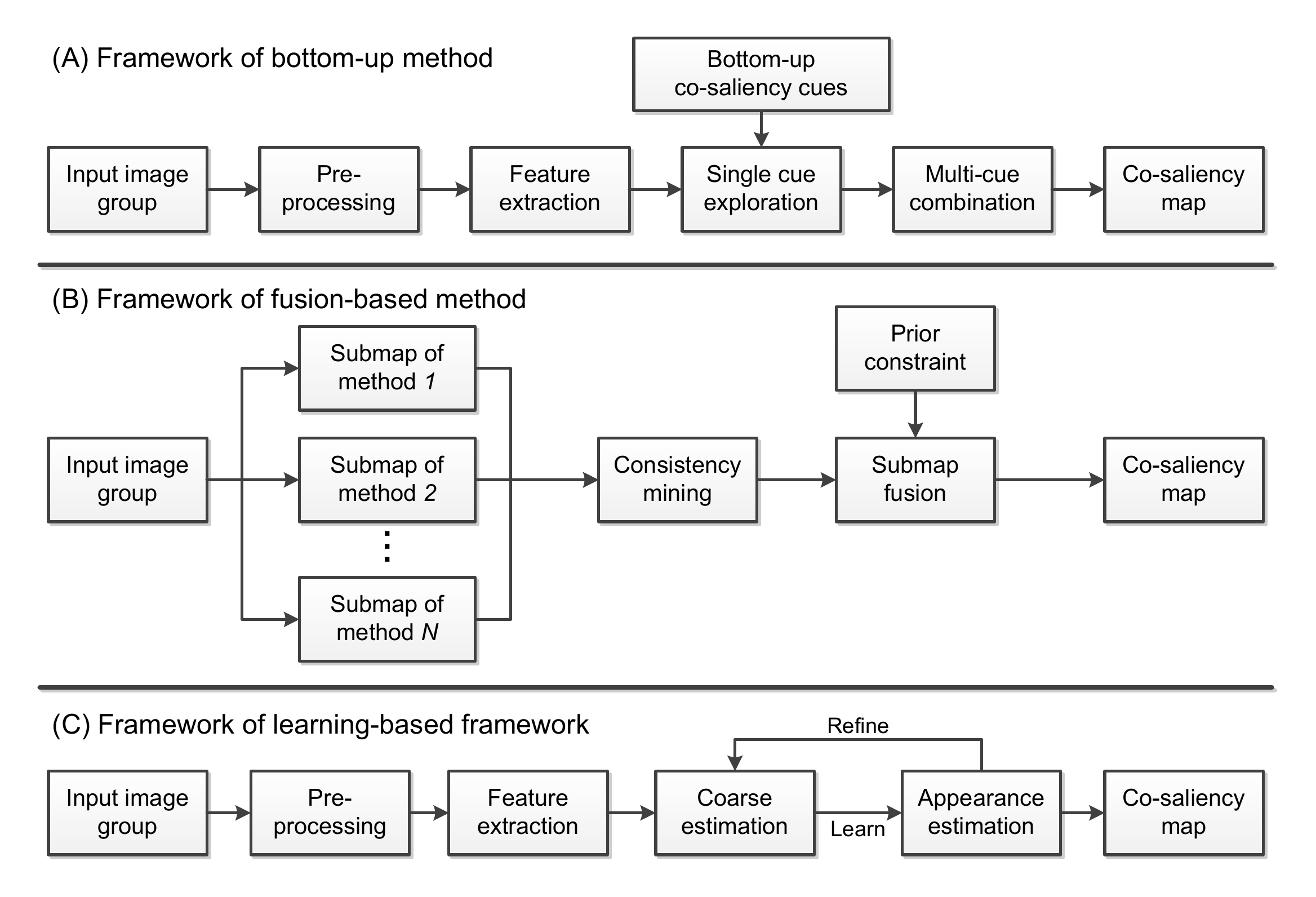}}
\caption{ {Co-saliency detection frameworks: (A) Bottom-up, (B) Fusion-based, and (C) Learning-based.}}
\label{fig_frame}
\end{figure*}

\subsection{Bottom-up methods}

A common co-saliency detection method is to score each pixel/region in the image group by using manually designed co-saliency cues; this is called the bottom-up method~\cite{Li2011,Fu2013,JING_IEICE_2015,Ge201669,Shen2016}. For example, Li and Ngan proposed to detect pair-wise co-saliency by exploring the single-image saliency and the multi-image co-saliency cues~\cite{Li2011}. Afterwards, they further improved their previous model~\cite{Li2011} by using multi-scale segmentation voting to explore the object property of the foreground regions to generate the intra-image saliency maps and extracting more powerful local descriptors to compute the inter-image saliency maps~\cite{Li2013}. Both of these two methods generated the final co-saliency maps by a weighted combination of the inter-image saliency maps and the intra-image saliency maps. Liu et al.~proposed a hierarchical segmentation based co-saliency model~\cite{Liu_SPL_2014}. They explored the local similarity and the object prior of each image region in the fine-level segments and coarse-level segments, respectively. Then, the global similarity was derived by simply summing the local similarities of several of the most similar segments from other images. Finally, co-salient objects were detected by fusing the local similarity, the global similarity, and the object prior.

Fig.~\ref{fig_frame} (A) presents the flowchart of bottom-up co-saliency detection methods, which mainly consist of the following stages: pre-processing, feature extraction, single cue exploration, and multi-cue combination. The pre-processing step segments the input image into many computational units ( {e.g. image pixels, pixel clusters, or superpixel segments}). Afterwards, features are extracted to explore the property of each computational unit based on each manually designed co-saliency bottom-up cue, i.e., the single cue exploration. Finally, results obtained from each bottom-up co-saliency cue are integrated together to generate the co-saliency maps for the input images. The key problem of bottom-up methods is the co-saliency cue. A basic role of the co-saliency cue is~\cite{Chang2011,Fu2013}:
\begin{equation}
\text{Co-saliency} = \text{Saliency} \times \text{Repeatedness}.
\end{equation}
For exploring the bottom-up cues, adopted techniques have changed from the early simple ones, e.g., matching and ranking, to the recent more powerful ones, e.g., matrix factorization and pattern mining. For the weighted combination, most bottom-up methods adopted simple strategies like multiplication or averaging, while some recent methods~\cite{Zhang_CVPR_2015,Li_SPL_2015} have applied better ways to take advantage of both multiplication and averaging.

 {One typical bottom-up method is the cluster-based co-saliency detection approach~\cite{Fu2013}, which measured the cluster-level co-saliency by using three bottom-up saliency cues. The global corresponding relation is built by clustering. For each cluster $t$, its co-saliency is formulated as:}
\begin{equation}\begin{split}
&\text{Co-Saliency}^t = \text{Cue}_{Con}^t \times \text{Cue}_{Spa}^t \times \text{Cue}_{Corr}^t, \\
\text{with:} &\\
&\text{Cue}_{Con}^t = \sum\nolimits_{\tau\neq t} \frac{n^\tau}{N} (||\textbf{u}^\tau-\textbf{u}^t||_2), \\
&\text{Cue}_{Spa}^t = \frac{1}{n^t} \sum\nolimits_{p \in C_t} \mathcal{N}(||\textbf{z}_p-\textbf{o}_p||^2|0,\sigma^2), \\
&\text{Cue}_{Corr}^t = 1/(\text{Var}(\textbf{q}^t)+1),
\end{split}\end{equation}
 {where $\text{Cue}_{Con}^t$, $\text{Cue}_{Spa}^t$, and $\text{Cue}_{Corr}^t$ indicate the contrast, spatial, and corresponding cues, respectively. $\textbf{u}^t$, $n^t$, and $\textbf{q}^t$ denote cluster center representation, the number of pixels, and the distribution of clusters $C_t$, respectively. The Gaussian kernel $\mathcal{N}(\cdot)$ computes the Euclidean distance between $\textbf{z}_p$ (the coordinate of the pixel $p$) and the coordinate of the corresponding image center $\textbf{o}_p$. The variance $\sigma^2$ is the normalized radius of images. $N$ denotes the pixel number of all images in the given image group. $\text{Var}(\textbf{q}^t)$ denotes the variance of the histogram $\textbf{q}^t$. More details can be found in~\cite{Fu2013}.}

As the most common form of co-saliency detection, bottom-up methods have obtained great development during the past few years. However, as these kinds of co-saliency detection approaches heavily rely on manually designed bottom-up cues pre-defined in their frameworks, they are typically too subjective and thus cannot generalize well to various scenarios encountered in practice.

\subsection{Fusion-based methods}

Rather than attempting to discover informative cues from the collection of multiple related images for representing co-salient objects, the fusion-based methods aim at mining the useful knowledge from the prediction results obtained by several existing saliency or co-saliency algorithms. They fuse those prediction results or the mined knowledge to generate the final co-saliency maps. For instance, Cao et al.~applied low-rank decomposition to exploit the relationship of generated maps from multiple existing saliency and co-saliency approaches to obtain the self-adaptive weights, and then used these weights to combine the maps for generating the final co-saliency map~\cite{Cao2013_ICME}.  A reconstruction-based fusion approach is proposed in~\cite{Cao_MM_2014}, which combines the knowledge mined from several existing saliency detection methods based on the reconstruction errors. Huang et al. fused the obtained multi-scale saliency maps by using low-rank analysis and introducing a GMM-based co-saliency prior~\cite{Huang_ICME_2015}.

Fig.~\ref{fig_frame} (B) illustrates the general flowchart of the fusion-based co-saliency detection methods. Specifically, first multiple coarse prediction submaps are obtained by applying existing saliency or co-saliency detection methods to images in the given image set. Then, the consistency property of the co-saliency regions are mined from the obtained coarse prediction submaps.  {Usually, some prior constraints (i.e., the helpful learning constraints or priorities) are also applied to infer the fusion weights of the coarse prediction submaps. For example, the methods in~\cite{Cao_TIP_2014,Cao_MM_2014,Huang_ICME_2015} mined the consistency property among the salient regions through the rank constraint and the reconstruction error, respectively.} Finally, the previously obtained prediction results are fused based on this guided knowledge to generate the final co-saliency maps by:
\begin{equation}
\text{Co-saliency} = \sum_i \text{Weight}_i \cdot \text{Submap}_i,
\end{equation}
where $Submap$ is the  coarse prediction submap, and $Weight$ is the guided fusion weighting based on consistency property and prior constraint.

 {One typical model of fusion-based method is the rank constraint-based self-adaptively weighting strategy~\cite{Cao_TIP_2014}. Given the feature matrix $\textbf{H}=[\textbf{H}^1,\textbf{H}^2,\cdots,\textbf{H}^K]^T \in \mathbb{R}^{KM\times D}$, where $\textbf{H}^k=[\textbf{h}_1^k,\textbf{h}_2^k,\cdots,\textbf{h}_M^k]^T \in \mathbb{R}^{M\times D}$, $K$, $M$, and $D$ indicate the number of images, the number of the extracted submaps for each image, and the dimension of the feature, respectively. The consistency among these submaps can be mined by solving a low-rank recovery problem~\cite{deng2013low}:}
\begin{equation}\begin{split}
&\arg\min_{\textbf{L},\textbf{E}}(||\textbf{L}||_*+\lambda ||\textbf{E}||_1), \\
&s.t. \ \ \ \ \textbf{H}=\textbf{L}+\textbf{E},
\end{split}\end{equation}
 {where $||\cdot||_*$ indicates nuclear norm and $||\cdot||_1$ is $\ell_1$-norm. The matrix $\textbf{E}$ is the error matrix between $\textbf{H}$ and the low rank matrix $\textbf{L}$, where $\textbf{E}^k=[\textbf{e}_1^k,\textbf{e}_2^k,\cdots,\textbf{e}_M^k]^T \in \mathbb{R}^{M\times D}$ is the error matrix of the $k$-th image. Then, the self-adaptive weight $w_j^k$ for the $j$-th submap of the $k$-th image can be obtained by:}
\begin{equation}
w_j^k=\frac{exp(\xi_j^k)}{\sum_{j=1}^M exp(\xi_j^k)}, \ \ \ \ \xi_j^k=-||\textbf{e}_j^k||_2.
\end{equation}
 {Finally, the co-saliency map is integrated with these multiple saliency cues by weighting the submaps self-adaptively. More details on this method can be found in~\cite{Cao_TIP_2014}.}

Generally, the fusion-based co-saliency detection methods usually achieve promising results as they can make further improvement based on multiple existing (co-)saliency detection approaches. Another advantage of these methods is that they can be flexibly embedded within various existing (co-)saliency detection methods, which makes them easier to adapt to different scenarios. However, as they heavily rely on the existing (co-)saliency detection methods, when most of the adopted (co-)saliency techniques only provide very limited or imprecise information regarding the co-salient objects, the final performance of the fusion-based co-saliency detection methods may also drift to the undesired image regions.

\subsection{Learning-based methods}

 {The third category of co-saliency detection techniques is the learning-based methods. These methods aim at directly learning the patterns of the co-salient objects from a given image group. Notice that although some fusion-based methods or bottom-up methods might also adopt machine learning techniques in their frameworks, they do not belong to this category as they only use the machine learning techniques to formulate certain computational blocks rather than learning the co-saliency patterns.} Specifically, Cheng et al.~\cite{Cheng_VC_2014} exploited correlations across Internet images within the same categories by first building coarse appearance models for target image regions and their backgrounds and then using the appearance models to improve saliency detection and image segmentation. Zhang et al.~\cite{Zhang_ICCV_2015,zhang2016PAMI} designed a self-paced multiple-instance learning (SP-MIL) framework to gradually learn the faithful patterns of the co-salient objects from confident image regions to ambiguous ones.  {As shown in Fig.~\ref{fig_frame} (C), the common property of these methods is that they first adopt off-the-shelf unsupervised saliency detection approaches to provide an initial coarse estimation. Then, the iterative self-learning schemes were designed to gradually incorporate top-down knowledge to learn to estimate the appearance of the co-salient object (by using either the generative model like GMM in~\cite{Cheng_VC_2014} or the discriminative model like SVM~\cite{chang2011libsvm}  in~\cite{Zhang_ICCV_2015,zhang2016PAMI}) and refine the obtained co-saliency maps.}

One typical model of learning-based method is self-paced multiple-instance learning (SP-MIL)~\cite{Zhang_ICCV_2015}. Given an image group containing $K_{+}$ images, they treated these images as the positive bags and searched $K_{-}$ similar images from other image groups as the negative bags. In each image, the extracted superpixels with their feature representation $\textbf{x}_{i}^{(k)}$ were treated as instances to be classified, where $k\in [1, K], K=K_{+}+K_{-}$ denotes the bag index of each image and   $i\in [1, n_{k}]$ denotes the instance index of each superpixel in the $k$-th bag.  The objective function of SP-MIL was formulated as:
\begin{equation}\begin{split}
& \underset{\textbf{w},b,\textbf{y}^{(1)},...,\textbf{y}^{(K_{+})},\textbf{v}\in [0,1]^{n}}{\min}
\mathbb{E}
(\textbf{w},b,\textbf{y}^{(1)},...,\textbf{y}^{(K_{+})},\textbf{v}) = \\
& \dfrac{1}{2}||\textbf{w}||_{2}^{2} + \sum_{k=1}^{K} \sum_{i=1}^{n_{k}} v_{i}^{(k)} L(y_{i}^{(k)},g(\textbf{x}_{i}^{(k)};\textbf{w},b)) + f(\textbf{v};\lambda,\gamma) \\
& s.t. \ \  ||\textbf{y}^{(k)}+1||_{0} \geq 1, k=1,2,...,K_{+},
\end{split}\end{equation}
where $\textbf{v}=[v_{1}^{(1)},...,v_{n_{1}}^{(1)}, v_{1}^{(2)},...,v_{n_{2}}^{(2)},...,v_{n_{K}}^{(K)}] \in \Re^{n}$ denotes the importance weights for the instances, $\textbf{y}^{(k)}=[y_{1}^{(k)},y_{2}^{(k)},...,y_{n_{k}}^{(k)}] \in \Re^{n_{k}}$ denotes the labels for instances in the $k$-th bag, $L(y_{i}^{(k)},g(\textbf{x}_{i}^{(k)};\textbf{w},b))$ denotes the hinge loss of $\textbf{x}_{i}^{(k)}$ under the linear SVM classifier $g(\textbf{x}_{i}^{(k)};\textbf{w},b)$ with the weight vector $\textbf{w}$ and bias parameter b. The constraint enforces the condition that at least one positive instance to be emerged in each positive bag. $f(\textbf{v};\lambda,\gamma)$ is the self-paced regularizer that introduces meaningful biases,  {e.g., the easiness bias and diversity bias which aid learning with confident and diverse training samples, respectively,} into the learning mechanism to better discover the patterns of co-salient objects. The regularizier may have many different forms.

In such classification-based co-saliency detection frameworks, most of the knowledge about the co-salient object regions are inferred by the designed learner automatically rather than heavily relying on some manually designed metrics as in other categories of co-saliency detection methods. Essentially, with an applicable learning strategy, the data-driven classification-based methods have the potential to capture better patterns of co-salient objects than the methods based on human-designed metrics. This because these hand-designed metrics are typically too subjective and cannot generalize well to various scenarios encountered in practice. The disadvantage of the classification-based co-salient detection methods is the execution time of the iterative training process, especially when the convergence condition is hard to achieve.

\section{ {Evaluation}}
\label{sec-evaluation}

\subsection{ {Co-saliency Datasets}}

 \begin{figure*}[!t]
 	\centerline{\includegraphics[width=\textwidth]{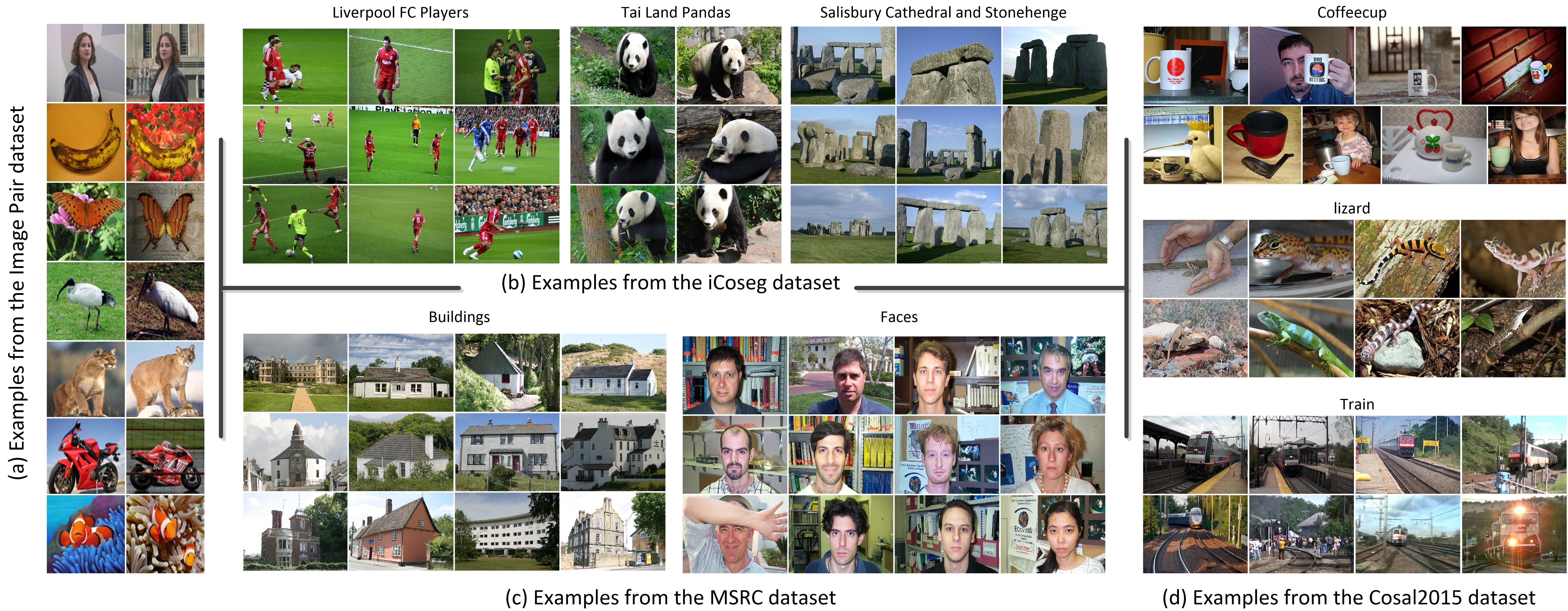}}
 	\caption{Some examples from the Image Pair, iCoseg, MSRC, and Cosal 2015 datasets.}
 	\label{img_dataset}
 \end{figure*}

\begin{table*}[!t]
\centering
\tiny
\caption{Brief illustration of datasets that are widely used in the evaluation of the co-saliency detection approaches. SAT is short for state-of-the-art.}
\begin{tabular}{lcccccccc}
	\hline
	{}         & image  & group  & group &     image      &         STA         &        Object        &    Background    &           Publish           \\
	{}         & number & number & size  &   resolution   &     performance     &       property       &     property     &            year             \\ \hline\hline
	Image Pair &  210   &  115   &   2   & 128$\times$100 & AUC: 0.97, AP: 0.93 &        Single        &      Clean       &     2011~\cite{Li2011}      \\
	iCoseg     &  643   &   38   & 4-42  & 500$\times$300 &  AP: 0.87, F: 0.81  &       Multiple       & Clean or complex & 2010~\cite{batra2010icoseg} \\
	MSRC       &  230   &   7    & 30-53 & 320$\times$210 &  AP: 0.85, F: 0.77  &       Complex        &    Cluttered     & 2005~\cite{winn2005object}  \\
	Cosal2015  &  2015  &   50   & 26-52 & 500$\times$333 &  AP: 0.74, F: 0.71  & Multiple and complex &    Cluttered     & 2016~\cite{Zhang_2016_IJCV} \\ \hline
\end{tabular}%
\label{Tab_dataset_Related}%
\end{table*}%

 {Co-saliency detection approaches are mainly evaluated on four public benchmark datasets: the Image Pair dataset~\cite{Li2011}, the iCoseg dataset~\cite{batra2010icoseg}, the MSRC dataset~\cite{winn2005object}, and Cosal2015 dataset~\cite{Zhang_2016_IJCV}. A brief summary of these datasets is shown in Table~\ref{Tab_dataset_Related}. These datasets are described and discussed in more detail next.}

 {The Image Pair dataset~\cite{Li2011} (see Fig.~\ref{img_dataset} (a)) contains 105 image pairs (i.e. 210 images) with resolution around 128$\times$100 pixels. It also has manually labeled pixel-level ground truth data and is the first benchmark dataset built for evaluating the performance of co-saliency detection. Each image pair contains one or more similar objects in different backgrounds.This dataset has been frequently utilized by the co-saliency detection approaches designed in early years (e.g.,~\cite{Chang2011,Chen2010,Li2011,Fu2013}). As the content of images in the Image Pair dataset is not as cluttered as other benchmark datasets, the state-of-the-art co-saliency detection methods have achieved promising performance (0.973 and 0.933 in terms of AUC and AP, respectively, according to Zhang et al.~\cite{Zhang_TNNLS_2015}) on it.}

 {The iCoseg dataset~\cite{batra2010icoseg} (see Fig.~\ref{img_dataset} (b)) is a large-scale publicly available dataset that is widely used for co-saliency detection. It consists of 38 image groups (643 images in total) along with pixel-level ground-truth hand annotations. Note that each image group in the iCoseg dataset contains 4 to 42 images rather than two images in the Image Pair dataset. Also, most images in the iCoseg dataset contain complex backgrounds and multiple co-salient objects. Thus, the iCoseg dataset is considered as a more challenging dataset for co-saliency detection and is widely used to evaluate the modern co-saliency detection techniques. As reported by Zhang et al.~\cite{Zhang_ICCV_2015}, the performances of the state-of-the-arts on this dataset are around 0.87 and 0.81 in terms of AP and F-measure, respectively. Since,
theoretically, the upper bound of AP and F-measure should be 1, there is still room for further improvement of co-saliency detection techniques on this dataset.}

 {Another dataset which is widely used for evaluating co-saliency detection approaches in recent time is the MSRC dataset~\cite{winn2005object}. The MSRC dataset consists of 8 image groups (240 images) with manually labeled pixel-wise ground truth data. As shown in Fig.~\ref{img_dataset} (c), image groups like airplane, car, and bicycle are contained in this dataset. Notice that the image group of `grass' is not usually used to evaluate co-saliency detection performance as it does not have co-salient objects in each image. Compared with the iCoseg dataset, there are different colors and shapes for the co-salient objects in image groups of the MSRC dataset, making it more challenging. Thus, most of the existing co-saliency detection approaches can only obtain performance less than 0.85 and 0.77 in terms of AP and F-measure, respectively.}

 {The Cosal2015 dataset~\cite{Zhang_2016_IJCV} (see Fig.~\ref{img_dataset} (d)) is a newly published dataset that is particularly established for the research in co-saliency detection. This dataset contains 50 image groups (2015 images in total) collected from the ILSVRC2014 detection benchmark~\cite{russakovsky2015imagenet} and the YouTube video set~\cite{prest2012learning}. As shown in Fig.~\ref{img_dataset} (d), the images in this dataset are highly cluttered. Moreover, in some cases, there are multiple salient objects in each single image, which, however, do not commonly appear in the image group. This causes a lot of ambiguity for the algorithms to determine which image regions are the co-salient ones. Consequently, the Cosal2015 dataset appears to be much bigger and more challenging than other existing datasets. Further, it provides pixel-level annotations manually labeled by 20 subjects. The state-of-the-art performance on this dataset only reaches to 0.74 and 0.71 in terms of AP and F-measure, respectively.}

\subsection{ {Evaluation Scores}}

 {To evaluate the performance of the co-saliency detection results, the existing literature has adopted six criteria including  precision-recall (PR) curve, average precision (AP) score, F-measure curve (F-score), mean F-score (mF), receive operator characteristic (ROC) curve, and AUC score, respectively.}

 {PR curve and AP score are generated by separating the pixels in a saliency map into salience or non-salience by varying the quantization threshold within the range [0, 1]. The resulting true positive rate (or the recall rate) versus precision rate at each threshold value forms the PR curve. The area under the PR curve is calculated as the AP score. Similarly, the ROC curve is generated based on the false positive rate and the true positive rate, and the AUC score is obtained by calculating the area under the ROC curve. Specifically, the precision and recall are defined as:}
\begin{equation}
Precision=\frac{TP}{TP+FP}, \; Recall = \frac{TP}{TP+FN},  \nonumber
\end{equation}
 {where $TP$, $FP$ and $FN$ denote the number of true positives, false positives and false negatives, respectively, under a binary segmentation using a certain thresholding on the co-saliency map. The corresponding F-score is calculated as:}
\begin{equation}
	\text{F-score}=\dfrac{(1+\beta^2) \cdot Precision \cdot Recall}{\beta^2 \cdot Precision + Recall} ,
\end{equation}
 {where $\beta^2=0.3$ as suggested in~\cite{achanta2009frequency}. The threshold value versus the corresponding F-score forms the F-measure curve and the mean value of the F-scores under all the thresholds is the mean F-score (mF).}

 {Among the aforementioned evaluation criteria, the PR curve, F-measure curve, and ROC curve tend to be more informative than the corresponding AP score, mean F-score, and AUC score. Specifically, the PR curve can reflect how precision changes with recall, F-measure curve can reflect how F-score changes with the varying thresholds, and the ROC curve can reflect how true positive rate (recall) changes with false positive rate. However, the AP score, mean F-score, and AUC score are more frequently used in comparing different co-saliency detection approaches as these criteria can provide concise and comprehensive evaluation of the compared approaches. In addition, when the area of the co-salient regions is similar with the area of the image background regions, all of these criteria can provide meaningful evaluation results. However, when the area of the co-salient regions and the area of the image background regions are very different, we recommend using the PR curve, F-measure curve, AP score, and mean F-score as the ROC curve and AUC score will be influenced by the imbalanced data~\cite{davis2006relationship}. Moreover, compared with the PR curve and AP score, the F-measure curve and mean F-score are more favored when the precision and recall have similar (and high) values.}

\subsection{ {Evaluation of Some Representative Models}}

\begin{figure*}[t]
\centerline{\includegraphics[width=\textwidth]{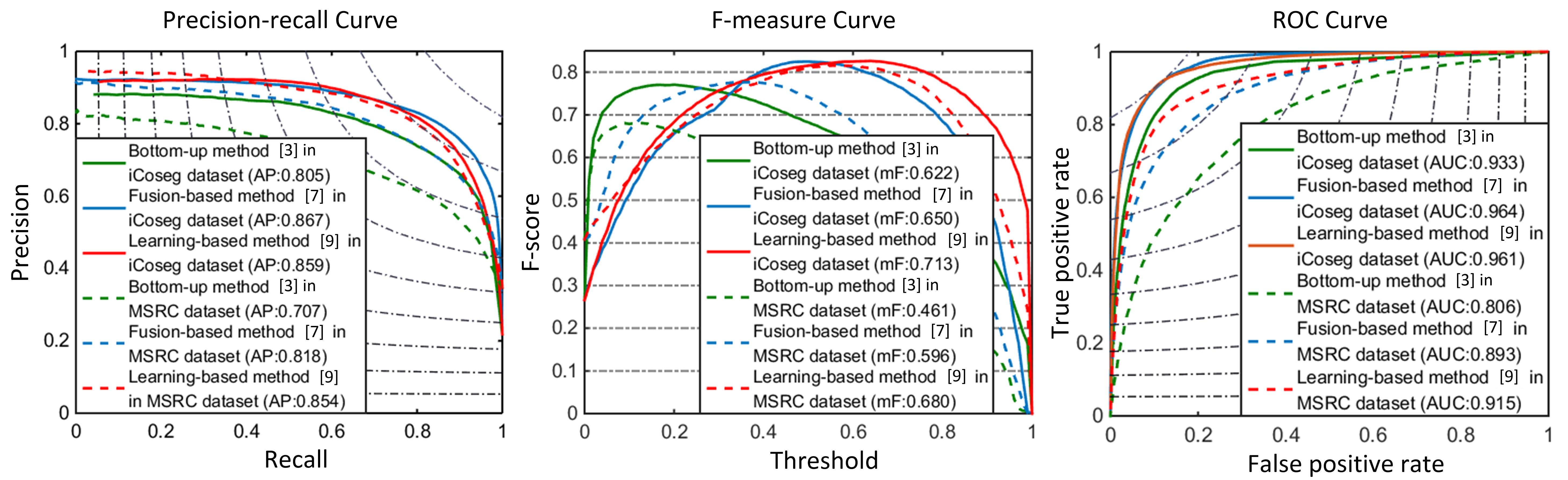}}
\caption{Comparison of the three categories of co-saliency detection techniques in terms of PR curve with AP score, F-measure curve with mF score, and ROC curve with AUC score, respectively.}
\label{img_curve}
\end{figure*}

\begin{figure*}[t]
\centerline{\includegraphics[width=\textwidth]{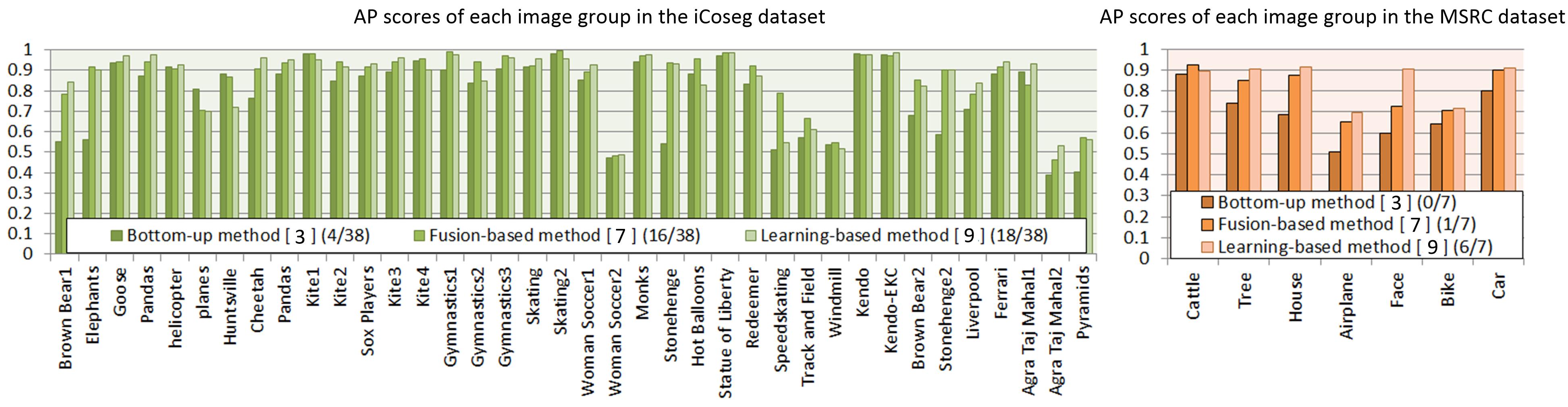}}
\caption{Comparison of the three categories of co-saliency detection technique on the image groups of the iCoseg and MSRC datasets, respectively. The fraction behind each category of methods reflects the successful cases, i.e., achieving superior performance than other methods, of the corresponding method.}
\label{exp_hist}
\end{figure*}

 {We compared three co-saliency detection methods on the iCoseg and MSRC datasets. These methods are Fu's method~\cite{Fu2013}, Cao's method~\cite{Cao_TIP_2014}, and Zhang's method~\cite{Zhang_ICCV_2015}, which, as described in Sec.~\ref{sec_algorithm}, are the representative approaches that have achieved state-of-the-art performance among the bottom-up methods, fusion-based methods, and learning-based methods, respectively. To make the experiment as fair as possible, we use the same feature representation as adopted in~\cite{Fu2013} for each method. Fig.~\ref{img_curve} shows the quantitative evaluation results in terms of the PR curve, AP score, F-measure curve, mF score, ROC curve, and AUC score, respectively. As can be seen, by taking better advantage of group-specific top-down priors, the fusion-based method and learning-based method generally outperform the bottom-up method. More specifically, as displayed in Fig.~\ref{exp_hist} we also compared the AP scores among these co-saliency detection methods in each image group of the iCoseg and MSRC datasets. It is easy to observe that the fusion-based method~\cite{Cao_TIP_2014} and the learning-based method~\cite{Zhang_ICCV_2015} obtain better performance on many more image groups than the bottom-up method~\cite{Fu2013}. This demonstrates that the top-down priors explored by the former two categories of co-saliency detection techniques can evidently improve the co-saliency detection performance in various cases.}

\section{ {Applications}} 
\label{sec_application}

 {In this section, we mention potential applications of co-saliency detection including object co-segmentation, video foreground detection, weakly supervised localization, image retrieval, multi-camera surveillance, and 3D object reconstruction.}

\subsection{ {Object co-segmentation}}

 {One of the most direct applications is object co-segmentation. It is a relatively high-level computer vision task which aims at generating binary masks to segment out the common objects from an image group~\cite{Zhu201612}. Implemented as a pre-processing step for the object co-segmentation task, co-saliency detection models can be applied to replace user interaction to provide the informative prior knowledge of visually similar objects with less supervision ~\cite{fu2015TIP,Fu2015_CVPR}. For example, Chang et al.~\cite{Chang2011} made one of the earliest efforts to propose a fully unsupervised approach to solve the problem of co-segmentation, where they established an MRF optimization model by introducing a co-saliency prior as the hint about possible foreground locations to replace user input and a novel global energy term to realize the co-segmentation process efficiently. In~\cite{Zhang_2016_IJCV}, Zhang et al.~adopted a self-adaptive threshold to segment the co-saliency maps generated by the proposed co-saliency model. The obtained binary co-segmentation masks are demonstrated to be competitive or even better than the state-of-the-art co-segmentation approaches.}

\subsection{ {Foreground discovery in video sequences}}

\begin{figure}[h]
\centerline{\includegraphics[width=8cm]{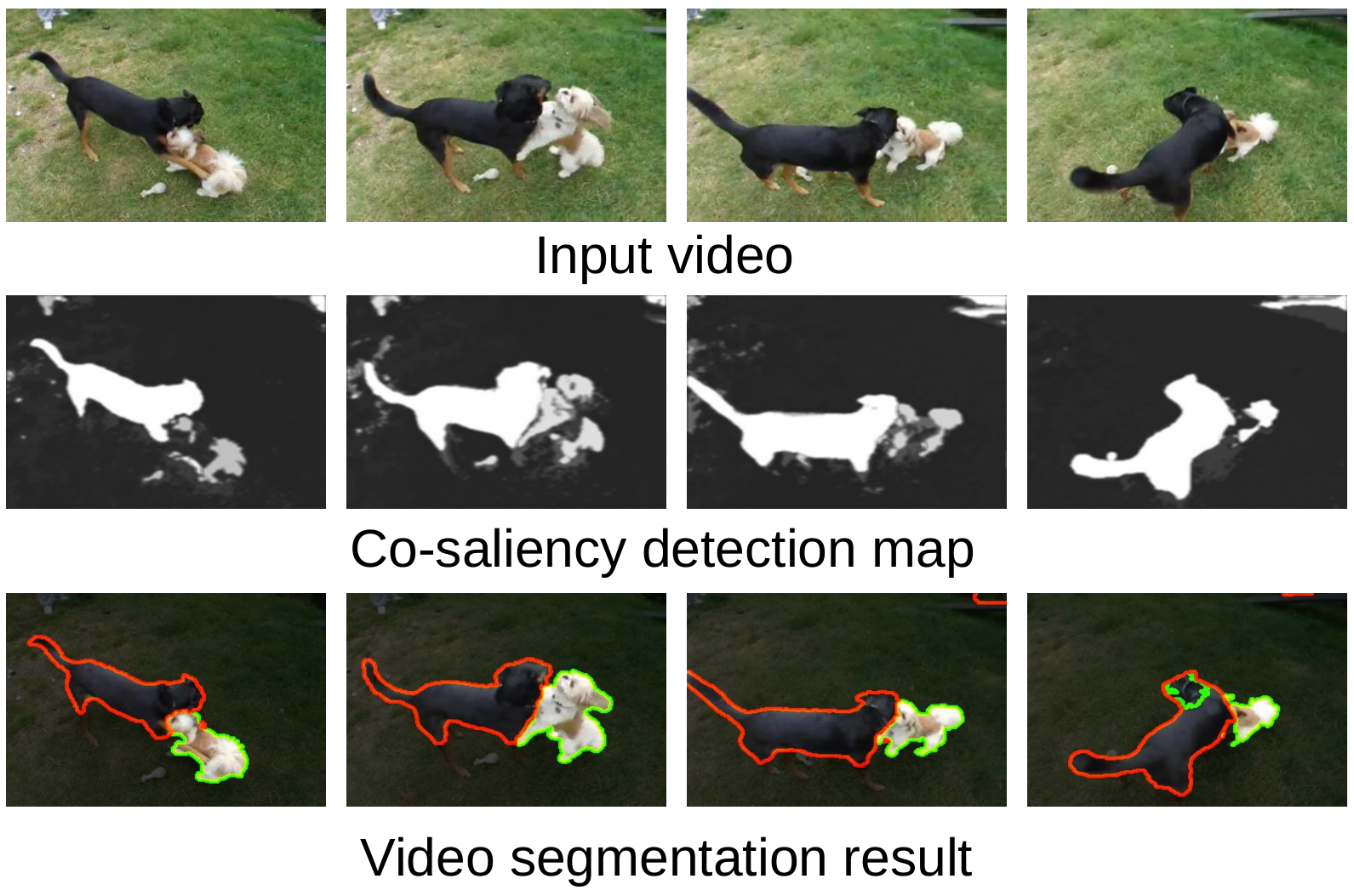}}
\caption{An example of video foreground segmentation based on co-saliency map~\cite{fu2015TIP}.}
\label{img_video_seg}
\end{figure}

 {With the goal of extracting foreground objects from video sequences, foreground discovery and segmentation in video is one potential application of the co-saliency detection techniques. It plays an important role in a large number of video analysis tasks~\cite{li2013survey,hong2011beyond,song2014unified} such as traffic monitoring, visual tracking, surveillance, and video summarization. It can discover the informative foreground components in the videos for improving the computational accuracy and efficiency. Essentially, foreground discovery has been formulated as a video saliency prediction problem in some previous works~\cite{Yang2015_TCSVT,40li2013exploring,Wang2015,Luo201645,ChangYXY15a}. Usually, the existing approaches compute video saliency by combining frame saliency cue and the temporal smooth cue among the consecutive frames to encode the information cues of contrast and motion~\cite{Mahadevan2010,Kim2011_TCSVT,Riche2013}. However, another global cue, i.e., the appearance consistency, can also be considered to model the foreground object. As mentioned in works~\cite{Fu2013,Cao2016,fu2015TIP}, the frames in video could be treated as a group of relative images, where a co-saliency detection technique is utilized to discover the common foregrounds, as shown in Fig.~\ref{img_video_seg}. Notice that in order to obtain better performance in this application, the co-saliency detection approaches should be further improved to handle the influence of the noisy images which may not contain the foreground object or action.}
 
\subsection{ {Weakly supervised localization}}

 {Weakly supervised localization (WSL) aims at learning object detectors to localize the corresponding objects by only using the image-level tags rather than the manually labeled bounding box annotations. It could implement more intelligent machines which are capable of automatically discovering the intrinsic patterns from the cheaply and massively collected weakly labeled visual data. As discussed in Sec.~\ref{sec_related_wsl}, co-saliency detection techniques can be applied to the task of WSL because without additional prior knowledge or supervision, salient and common cues are the two major useful pieces of information for initializing the WSL frameworks and they are what co-saliency detection models seek to figure out. Even though co-saliency detection methods can be directly applied to initialize the common objects in the positive images with weak labels~\cite{WSLannotation}, some informative knowledge from the negative images are beyond exploration, which may lead to sub-optimal solutions. Specifically, most co-saliency detection approaches detect co-salient objects by only exploring each given image group that contains a certain type of common objects (i.e., the positive images in the task of weakly supervised object localization), while the information from image groups weakly labeled as containing other types of objects (i.e., the negative images in the task of weakly supervised object localization) is not involved in the frameworks. Without such information, the weakly supervised learning process will lack the discriminative capability in localizing different types of objects.}

 {One way to solve this problem and thus further improve the performance when applying co-saliency detection technique in WSL is to involve the prior information contained in the negative images in the formulation of co-saliency detection. For example, Zhang et al.~formulated co-saliency in a Bayesian framework~\cite{Zhang_CVPR_2015} as:}
\begin{equation}\begin{split}
\text{Cosal}(x_{m,p}) & =\text{Pr}(y_{m,p}=1|x_{m,p})\\
&\propto\underbrace{\dfrac{1}{\text{Pr}(x_{m,p})}}_{Intra-image \atop contrast}\underbrace{\text{Pr}(x_{m,p}|y_{m,p}=1)}_{Inter-image \atop consistency}\\
\end{split}\end{equation}
 {where $x_{m,p}$ denotes the feature representation of the $p$-th superpixel region in the $m$-th image, and $y_{m,p}$ is the label of $x_{m,p}$, indicating whether $x_{m,p}$ belongs to the co-salient region. {Intra-image contrast indicates the contrast within each single image while inter-image consistency indicates the consistency within the image group.} As can be seen, when applying the above formulation in WSL, it can only consider the intra-image contrast and the inter-image consistency in positive images. To further leverage the useful information contained in the negative images, one can follow the Bayesian rule to extend the formulation as:}
\begin{equation}\begin{split}
	\text{Cosal}(x_{m,p}) & =\text{Pr}(y_{m,p}=1|x_{m,p})\\
	&\propto\underbrace{\dfrac{1}{\text{Pr}(x_{m,p})}}_{Intra-image \atop contrast}[\underbrace{\text{Pr}(y_{m,p}=1)}_{Foreground \atop prior} \underbrace{\text{Pr}(x_{m,p}|y_{m,p}=1)}_{Inter-image \atop consistency}\\
	& \ \ \ \ -\underbrace{\text{Pr}(y_{m,p}=0)}_{Background \atop prior} \underbrace{\text{Pr}(x_{m,p}|y_{m,p}=0)}_{Inter-group \atop seperability}]
\end{split}\end{equation}
 {The extended formulation introduces the factor of inter-group separability as well as another two priors, which can be used to localize the objects in the weakly labeled images effectively.}

\subsection{ {Image retrieval}}

 {An image retrieval system is a computer system for browsing, searching and retrieving images from a large database of digital images. In such a system, a user may provide query images and the system will return images ``similar'' to the query. Recently, the object-sensitive image pair-distance (i.e., the image pair-distance that mainly focuses on the similarity of the objects contained by the image pairs) has been demonstrated to be beneficial for image retrieval systems, especially the ones based on global feature comparison~\cite{yang2011object,Papushoy2015156,Cheng_VC_2014}. Thus, as a promising way to generate robust image pair-distances, co-saliency detection techniques can also be applied to the image retrieval task.}

 {For example, Fu et al.~has provided an efficient and general robust image pair-distance based on the co-saliency maps~\cite{Fu2013}. Given an image pair $\{I_{i}\}_{i=1}^{2}$, each image can be segmented into the co-salient foreground regions $I^{f}$ and the background regions $I^{b}$ by thresholding the co-saliency maps. Then, the co-saliency-based image-pair distance $D(.)$ was defined as:}
\begin{equation}
	D(I_1, I_2) = w_f \cdot Dist(I_{1}^{f},I_{2}^{f}) + w_b \cdot Dist(I_{1}^{b},I_{2}^{b}),
\end{equation}
 {where $w_f$ and $w_b$ are the foreground and background weights, respectively, with $w_f + w_b = 1$. $Dist(.)$ denotes the traditional image distance. As can be seen, this equation makes the co-salient object regions play a more important role in the image-pair distance computing.}

\subsection{ {Multi-camera surveillance}}

\begin{figure}[h]
	\centerline{\includegraphics[width=8cm]{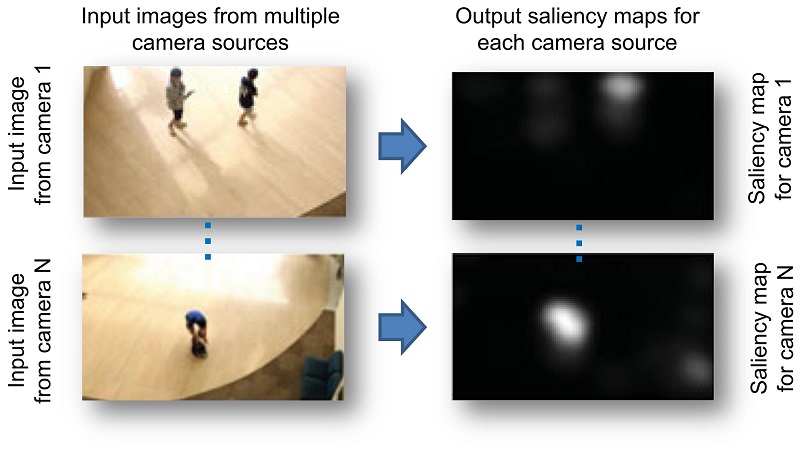}}
	\caption{Illustration of multi-camera saliency.}
	\label{figure8}
\end{figure}

 {Video surveillance in a large and complex environment always requires the use of multiple cameras~\cite{kettnaker1999bayesian}. This area has led to a large research interest in multi-camera surveillance. In multi-camera surveillance, one important task is to automatically detect and track the person with abnormal (distinct) behaviors monitored by multiple camera sources. Recently, Luo et al.~posed this problem as the multi-camera saliency estimation (shown in Fig.~\ref{figure8}), which extended the conventional saliency detection methods to identify the regions of interest with information from multiple camera sources in an integrated way~\cite{luomulti}. As can be seen, the problem of multi-camera saliency estimation is highly related to co-saliency detection. Both of these two tasks need to discover common and salient image regions in multiple related images. Thus, co-saliency detection techniques can also be applied to the task of multi-camera saliency and multi-camera surveillance. It should also be noted that the images used in multi-camera surveillance are usually captured by the cameras positioned in various locations and thus contain huge diversity in terms of visual perspective, which should be considered additionally when applying the co-saliency detection technique in multi-camera surveillance.}
 
\subsection{ {3D object reconstruction from 2D images}}

 {3D object reconstruction is a core problem in the computer vision community and has been extensively researched over the last few decades. It has been largely solved in the multi-view rigid case, where calibration and correspondences can be estimated. Thus it is a well-understood geometric optimization problem. More recent interests are on class-based 3D object reconstruction~\cite{vicente2014reconstructing,kar2014category}, where the goal is to reconstruct the objects belonging to the same category and pictured in each single 2D image of a given image group. Specifically, Vicente et al.~proposed to first estimate camera viewpoint of each image using rigid structure-from-motion and then reconstruct object shapes by optimizing over visual hull proposals guided by loose within-class shape similarity assumptions~\cite{vicente2014reconstructing}. Kar et al.~proposed the deformable 3D models which can further implement 3D object reconstruction in a more unconstrained setting~\cite{kar2014category}. However, even for these two most recent approaches, both of them need manually labeled ground-truth masks for the objects appearing in each image, which requires tedious human labeling and limits the capability to scale up for reconstructing 3D objects in arbitrary categories.}

 {Fortunately, due to their rapid development in recent years, co-saliency detection techniques can be used to alleviate the tedious human labeling task and scale up the 3D object reconstruction for any given object category. As shown in Fig.~\ref{figure9}, for a given object category, we can first search for the relevant images from the Web, where many diverse views of the objects in same category are available. Some processes can be applied to purify the searched images of noisy data. Then, any off-the-shelf co-saliency detection technique can be applied to detect the co-salient objects appearing in the image collection. Afterwards, some segmentation techniques, such as direct shareholding and graph cut, can be adopted to generate the masks for the objects. Finally, by estimating the camera viewpoints of each image, the 3D object models can be reconstructed from the obtained co-saliency masks of each 2D image.}

\section{Discussion}
\label{sec_discussion}

\subsection{Low-level features vs. High-level features}
\label{sec-feature}

As a basic yet critical component, the features adopted to represent the image pixels or regions significantly affect the performance of co-saliency detection algorithms. From the existing literatures, we observe that the early co-saliency detection methods are mainly based on low-level features. {For example, Li et al.~\cite{Li2011} described the image region appearance from the aspect of color variations and texture pattern. The color features were extracted from the RGB, Lab and YCbCr color spaces and the texture features were created
by combining a series of histograms of patchwords, which characterize the patch types occurring in the given images. Liu et al. ~\cite{Liu_SPL_2014} uniformly quantized each color
channel in the Lab color space into a number of bins to obtain a normalized global color histogram to represent each image region. Fu et al.~\cite{Fu2013} employed the CIE Lab color and Gabor filter responses with eight orientations to represent each image pixel.} These methods assumed that the common objects appearing in a given image group should have high consistency in terms of the low-level features and can be distinct from the image background based on these features.  {In contrast, some of the most recent works have tried to utilize high-level features and have obtained encouraging performance. Specifically, Zhang et al.~\cite{Zhang_2016_IJCV} built a higher-level representation for each object proposal bounding box via a domain adaptive Convolutional Neural Network (CNN), which equipped the conventional CNN architecture with additional transfer Restricted Boltzmann Machine (RBM) layers. In~\cite{zhang2016PAMI}, a hypercolumn feature representation, which was a combination of feature maps from different CNN layers, was extracted to represent each image superpixel region. The hypercolum feature representation captures both low-level and and high-level representations.

 \begin{figure*}[t]
	\centerline{\includegraphics[width=\textwidth]{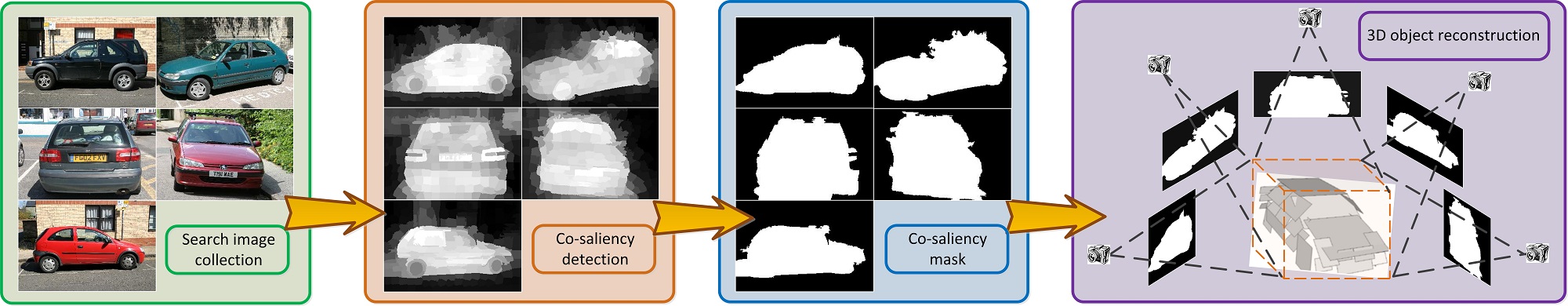}}
	\caption{Applying co-saliency detection for 2D image based 3D reconstruction.}
	\label{figure9}
\end{figure*}

Different from the low-level features, high-level features usually embed semantic information and thus can overcome the instability issue caused by the variations in viewpoints, shapes, and luminance. Consequently, they can model the concept-level properties of the co-salient objects more effectively. Notice that, in general, bottom-up co-saliency detection mechanisms are determined by conspicuity of stimulus from the environment. They highlight salient regions that are distinctive with respect to their surrounding context and have large consistency with object regions in other relevant images. Meanwhile, top-down mechanisms need to account for the endogenous cues that play a critical role in simulating situations that humans have specific tasks in mind.


\begin{figure*}[t]
\centerline{\includegraphics[width=\textwidth]{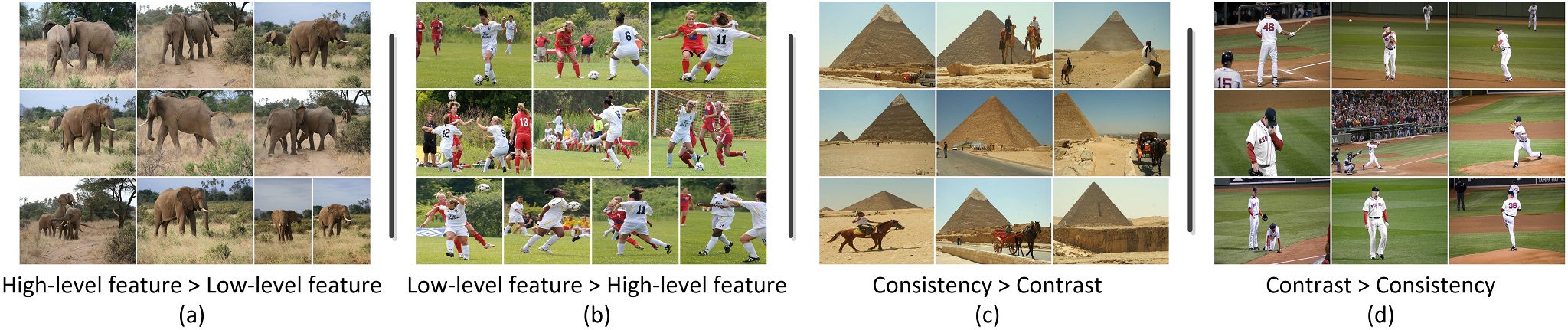}}
\caption{Some examples to illustrate the different co-saliency detection cases. (a,b) Intra-image contrast vs. Inter-image consistency, and (c,d) Low-level feature vs.~High-level features.}
\label{img_discuss}
\end{figure*}

Essentially, neither low-level nor high-level features can individually handle all the cases in co-saliency detection. As shown in Fig.~\ref{img_discuss} (A), the elephants in the image groups have the similar color and texture with the background regions containing the grass. However, at the semantic level, the image regions of elephants are related to the semantics like animals whereas the grass regions are related to the semantics like ground. Thus, high-level features could be used to detect co-saliency more efficiently in this case. On the contrary, as shown in Fig.~\ref{img_discuss} (B), the red football players and the white ones share the similar semantics for humans. However, as only the white players appear in every image of the given image group, the desirable co-salient regions should be the white players rather than the red ones. In this case, we need to adopt the low-level color feature to separate the real co-salient regions from the confusing ones.

Consequently, the proper features that should be used in co-saliency detection are hard to select and design because they are closely related to the content of a given image group. As there is no existing literature working on the idea of learning to select proper features according to the specific content of a given image group, we believe this is a promising yet challenging future direction.

\subsection{ {Intra-image contrast vs. Inter-image consistency}}

Intra-image contrast and inter-image consistency are the two most critical properties of co-salient object regions. {One direct way to gain insight regarding relative importance of these factors in co-saliency detection is based on the corresponding weights assigned to the intra-image contrast term and the inter-image consistency term. In this way, we can observe the assumptions or experimental evidences from the model settings and experiments of the existing literature.} For example, Li et al.~used equal weights to add the two factors to obtain the final estimation~\cite{Li2011}. Similarly, Du et al.~integrated these two factors via multiplication with equal weights~\cite{Du_SPL_2015}. These works indicate that the intra-image contrast and inter-image consistency have the same importance during co-saliency detection. Whereas Zhang et al. verified the parameter settings in their experiments and concluded that the inter-image consistency (with the weight of 0.7) is more important than the intra-image contrast (with the weight of 0.3)~\cite{Zhang_TNNLS_2015}.

Essentially, the importance of these two factors is related to the designed co-saliency detection algorithms and, more importantly, is highly related to the content of the image group. For example, as shown in Fig.~\ref{img_discuss} (C), the inter-image consistency is more important in discovering the co-salient objects in the image group of `Egypt Pyramids' whereas the intra-image contrast misleads the detector to focus on the regions of human and horse. On the contrary, for the image group of `Sox Players' in Fig.~\ref{img_discuss} (D), the inter-image consistency tends to be less important than the intra-image contrast as both the foreground and background regions would obtain high consistency values in such image group, which reduces the capability of the inter-image consistency to discover the co-salient object regions.

Consequently, one promising solution for assigning proper importance weights to these investigated factors in one algorithm would be to do so in a content-aware manner. This means that the designed algorithms should have the capability to automatically infer the proper weights for the factors based on the content of the specific image group. Along this direction, novel co-saliency detection methods are needed to further improve the detection performance.

\subsection{Bottom-up frameworks vs. Top-down frameworks}

Inspired by the conventional saliency detection methods, most of the existing co-saliency detection techniques are designed in a bottom-up manner, where the frameworks are established highly based on the human designed features and metrics to explore the visual stimulus in each image without any task-related guidance. Although it has been successfully used in the conventional saliency detection, performing co-saliency detection in such a bottom-up manner cannot obtain satisfactory results in many scenarios. The fundamental reason is that conventional saliency detection only needs to simulate the early stages of the visual attention system when humans are receiving the visual stimulus from one single image, while co-saliency detection needs to additionally formulate the abstract knowledge obtained during the human reviewing of the images in a given group. Thus, compared with the conventional saliency detection, co-saliency detection tends to be more complicated because only exploring the bottom-up cues is not enough\footnote{ {As mentioned in~\cite{borji2013state,itti1998model}, the regular saliency is often considered in the context of bottom-up computation because it intuitively characterizes some parts of a scene that appear to an observer to stand out relative to their neighbors. Such parts are exogenous stimulus like objects or regions. Top-down factors could also benefit visual attention modeling that is guided by endogenous stimulus to deal with situations that a human has specific tasks in mind. However, since much less is known about top-down attention mechanisms, these cues are considered to fall beyond the scope of regular saliency covered in this paper.}}.

Essentially, as discussed in the above two subsections, co-saliency detection is highly related to the content of a given image groups. In other words, co-saliency detection results would be largely influenced by the global-level information among the entire image group. Thus, a more proper way to formulate co-saliency should be in the top-down manner, where the global-level information could provide critical top-down guidance and priors for exploring the co-salient object regions in a specific image group. To introduce the top-down priors in co-saliency detection, some of the recent techniques have made efforts in two ways. The first way is to fuse multiple bottom-up (co-)saliency detection results under the guidance of the global-level information in a specific image group, i.e., the fusion-based approaches. The second way is to learn to model co-saliency under weak supervision provided by the image-level labels among different image groups, i.e., the (self-)learning-based methods. By leveraging the top-down priors, the fusion-based methods and learning-based methods can outperform the bottom-up methods in most cases.

\subsection{ {Image co-saliency vs. Video co-saliency}}

 {The majority of the co-saliency detection approaches reviewed in this paper are image-based. Some video co-saliency detection approaches have emerged most recently~\cite{xie2016video,Jerripothula2016}. Different from the image co-saliency that is performed on each given image group to discover common and salient patterns, video co-saliency aims at detecting co-salient objects in multiple videos by integrating the temporal coherence in each video with the co-occurrence of objects across multiple videos~\cite{xie2016video}. Its ultimate goal is to generate co-saliency maps for each video frame of a given video collection. Thus, compared with image co-saliency, video co-saliency needs to additionally consider information cues such as intra-video temporal saliency~\cite{xie2016video}, inter-video co-saliency~\cite{Jerripothula2016}, and intra-video co-saliency~\cite{Jerripothula2016}. Specifically, intra-video temporal saliency is generated by exploring the motion distinctiveness from background regions and temporal coherence in a period of consecutive frames. Inter-video co-saliency exploits the common patterns from different videos of similar objects to discover the common objects residing in different videos. Conversely, intra-video co-saliency is adopted to highlight the co-salient objects from the video frames within a single video with diverse backgrounds.}

 {From the above discussion, we can see that video co-saliency is an extension of image co-saliency. It tends to be more complex as it needs to explore the useful information from the motion modality and common patterns among different videos. In addition, it would have higher computational complexity requirements. This is a drawback since algorithms with high computational complexity can hardly scale up to large-scale video collections.}

\subsection{ {Deep learning for co-saliency detection}}

 { Deep learning techniques have recently been employed in a broad spectrum of applications in computer vision from  low-level computer vision tasks such as image super-resolution to high-level computer vision tasks such as action recognition~\cite{zhu2015fusing}. They achieve exceptional power and flexibility by learning to represent the task through a hierarchy of layers rather than relying on handcrafted features or hard-coded knowledge. In light of their superior capability, some of the most recent works have also adopted deep learning for addressing the co-saliency detection problem. Among these works, the first effort was made by Zhang et al.~\cite{Zhang_CVPR_2015,Zhang_2016_IJCV}\footnote{ {\cite{Zhang_2016_IJCV} is the journal extension version of~\cite{Zhang_CVPR_2015}.}}, who proposed to look deep to transfer higher-level representations by using a CNN with additional adaptive layers to better reflect the properties of the co-salient objects, especially their consistency in an image group. The experimental results reported by them demonstrated that using the higher-level representations extracted by the proposed CNN can lead to around 7\% performance gain. Different from~\cite{Zhang_CVPR_2015,Zhang_2016_IJCV}, the stacked denoising autoencoder, which is another kind of deep learning model, was adopted in~\cite{Zhang_TNNLS_2015} to transfer the saliency prior knowledge to estimate the intra-image contrast and mine intrinsic and general hidden patterns to discover the homogeneity of co-salient objects, respectively. More recently, by considering both low-level and the high-level feature representations, \cite{Zhang_ICCV_2015,zhang2016PAMI} proposed to build a hypercolumn feature representation by combining the CNN feature maps from shallow layers to deep layers. }

 {Existing deep learning-based co-saliency detection approaches utilize two important strategies: 1) transferring feature representation or helpful knowledge from the auxiliary source domain, and 2) mining the intrinsic common patterns of the co-salient objects appearing in the given image group. This is mainly due to the fact that deep learning techniques have superior capability in learning feature representation and mining hidden patterns, which happen to be two of the most critical factors for addressing the co-saliency detection problem. In addition, we also observe that the deep learning techniques used in the existing co-saliency detection approaches are performed in transfer learning or unsupervised learning manner, rather than the most commonly used supervised learning manner. This is because the co-saliency detection algorithms are performed without the supervision provided by the human annotation. Thus, most of the parameters in the adopted deep learning models are only pre-trained over an auxiliary dataset (e.g. the ImageNet dataset~\cite{russakovsky2015imagenet}) without any fine-tuning process on the task domain. This would limit the representation capability of the deep learning models. To address this problem, future work should focus on how to leverage the top-down information of a given image group to guide the fine-tuning process of the deep learning models and facilitate the end-to-end self-learning framework.}

\subsection{Challenges}

Although the co-saliency detection community has achieved substantial progress in the several past years, there is still a long road towards making co-saliency widely applicable to more practical applications. This is more related to the limitations of existing methods in terms of robustness and efficiency. To be more specific, we summarize the challenges that need to be addressed as follows.

The first challenge is the complexity of the image. For example, when the appearances (e.g. color, texture) of non-salient background and salient areas are similar, as shown in Fig.~\ref{fig_fail} (A), it is difficult to distinguish the foreground from the complex background solely based on the human visual fixation. A solution for this case is to employ more high-level features and to explore the semantic context. The discriminative features could help distinguish the foreground from the complex background.

\begin{figure}[h]
	\centerline{\includegraphics[width=8cm]{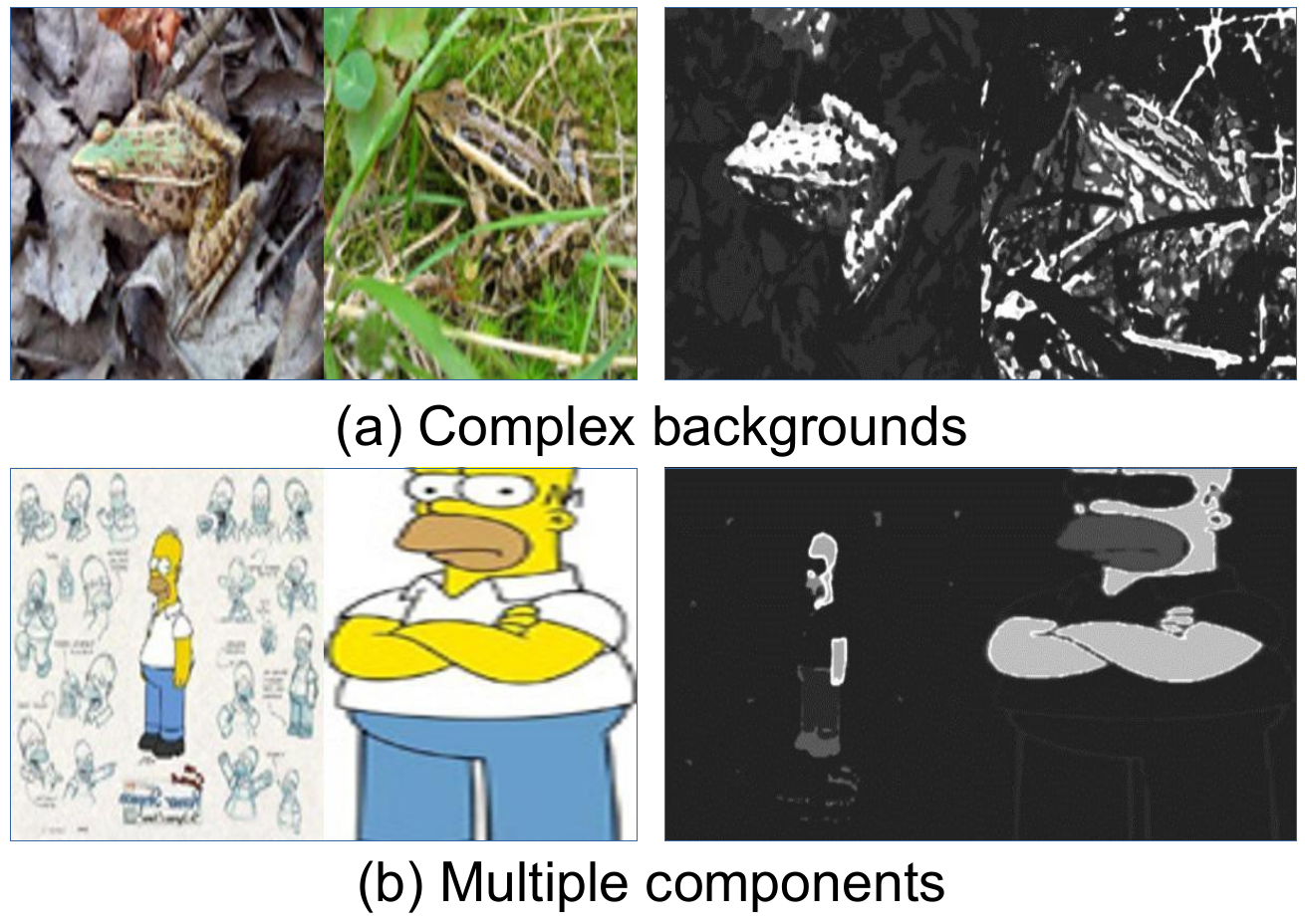}}
	\caption{Some challenging cases for co-saliency detection~\cite{Fu2013}.}
	\label{fig_fail}
\end{figure}

Another challenge is that if the foreground is composed of multiple components, as shown in Fig.~\ref{fig_fail} (B), the co-saliency detection method could not provide the entire object-level foreground. The main reason is that most co-saliency detection methods are based on a bottom-up role without heavy learning, which could not provide the high-level constraint. A solution for this case is to employ more high-level features and to explore the semantic context. One useful solution is to introduce the objectness constraint, which measures the likelihood of a region being an object. The objectness is based on general object properties and is widely employed to generate object bounding boxes and proposals. The objectness preserves the wholeness of the general object and could be used to segment the entire extend of the foreground.

The third challenge is the large-scale data. Co-saliency detection aims to distinguish the common foreground from multiple images, which is easily considered to apply to pattern discovery from large-scale data. Most co-saliency detection algorithms work well over small- and middle-sized datasets (within 100 $\sim$ 1000 images), but they do not easily deal with large-scale data. There are some unsolved issues, such as outliers and noise images, intra-class variation, and so on. How to deal with such challenging issues in large-scale data is a direction for future work.

 {The fourth challenge regards the efficiency of the co-saliency detection algorithms. Most co-saliency detection approaches mainly consist of the components of feature extraction, exploring the contrast information within each single image, and exploring the consistency information from multiple images. The feature extraction and exploration of the contrast information for each single image are independent with other images from the same group. Thus, parallel computation can be applied to efficiently complete these two components for large-scale image groups. However, the computational cost of exploring the consistency information from multiple images will increase dramatically with the increase in the scale of the image data, and this component will be the most time consuming component in co-saliency detection. To reduce the computation cost in exploring consistency information from multiple images, one direction is to establish more efficient computational frameworks. Along this direction, one can propose effective approaches to decompose a given large-scale image group into multiple smaller subgroups that have stronger intra-group correspondence and then implement co-saliency detection in each subgroup. Another direction is to design more efficient learning models with fast large-scale optimization algorithms. Along this direction, one can build models with online learning schemes or training sample selection mechanisms, and use computational physics, scientific computing, and computational geometry to design fast approximate algorithms.} 

\section{Conclusion}
\label{sec_conclusion}

In this paper, we reviewed the co-saliency detection literature. We described the recent progress in this field, analyzed the major algorithms, and discussed challenges and potential applications of co-saliency detection. In summary, co-saliency detection is a newly emerging and
rapidly growing research area in the computer vision community. It has derived from conventional saliency detection but deals with detecting and segmenting common salient objects in several images. Since a sheer number of images, often several images about one topic or location, is accessible from the Web, we believe that co-saliency detection techniques
hold a great promise in real-world applications. This motivated us to review the co-saliency detection techniques and discuss their fundamentals, applications, and challenges. We believe that this review is beneficial to researchers in this field as well as researchers working in other relevant areas, and will hopefully encourage more future works in this direction.

\bibliographystyle{ACM-Reference-Format}
\bibliography{Cosal_bib}

\end{document}